\documentclass[acmtog]{acmart}
\acmSubmissionID{553}

\usepackage[ruled]{algorithm2e} 
\usepackage{algorithmic}
\usepackage{subcaption}
\usepackage{graphicx}
\usepackage{multirow}

\SetAlFnt{\small}
\SetAlCapFnt{\small}
\SetAlCapNameFnt{\small}
\SetAlCapHSkip{0pt}

\copyrightyear{2026}
\acmYear{2026}
\setcopyright{cc}
\setcctype{by}
\acmConference[SIGGRAPH Conference Papers '26]{Special Interest Group on Computer Graphics and Interactive Techniques Conference Conference Papers}{July 19--23, 2026}{Los Angeles, CA, USA}
\acmBooktitle{Special Interest Group on Computer Graphics and Interactive Techniques Conference Conference Papers (SIGGRAPH Conference Papers '26), July 19--23, 2026, Los Angeles, CA, USA}
\acmDOI{10.1145/3799902.3811099}
\acmISBN{979-8-4007-2554-8/2026/07}

\newcommand{\comments}[1]{}
\begin{document}
\title{Sparse-to-Complete: From Sparse Image Captures to Complete 3D Scenes}

\author{Yiyang Shen}
\email{shenyiyang114@gmail.com}
\orcid{0000-0003-2311-0950}
\affiliation{%
  \institution{State Key Lab of CAD\&CG, Zhejiang University}
  \city{Hangzhou}
  \country{China}}

\author{Yin Yang}
\email{yangzzzy@gmail.com}
\orcid{0000-0001-7645-5931}
\affiliation{%
  \institution{University of Utah}
  \city{Salt Lake City}
  \country{USA}}

\author{Kun Zhou}
\email{kunzhou@acm.org}
\orcid{0000-0003-4243-6112}
\affiliation{%
  \institution{State Key Lab of CAD\&CG, Zhejiang University}
  \city{Hangzhou}
  \country{China}}
\affiliation{%
  \institution{Hangzhou Research Institute of Holographic and AI Technology}
  \city{Hangzhou}
  \country{China}}

\author{Tianjia Shao}
\authornote{Corresponding author}
\email{tjshao@zju.edu.cn}
\orcid{0000-0001-5485-3752}
\affiliation{%
  \institution{State Key Lab of CAD\&CG, Zhejiang University}
  \city{Hangzhou}
  \country{China}}
\affiliation{%
  \institution{Hangzhou Research Institute of Holographic and AI Technology}
  \city{Hangzhou}
  \country{China}}

\begin{teaserfigure}
  \centering
  \includegraphics[width=\textwidth]{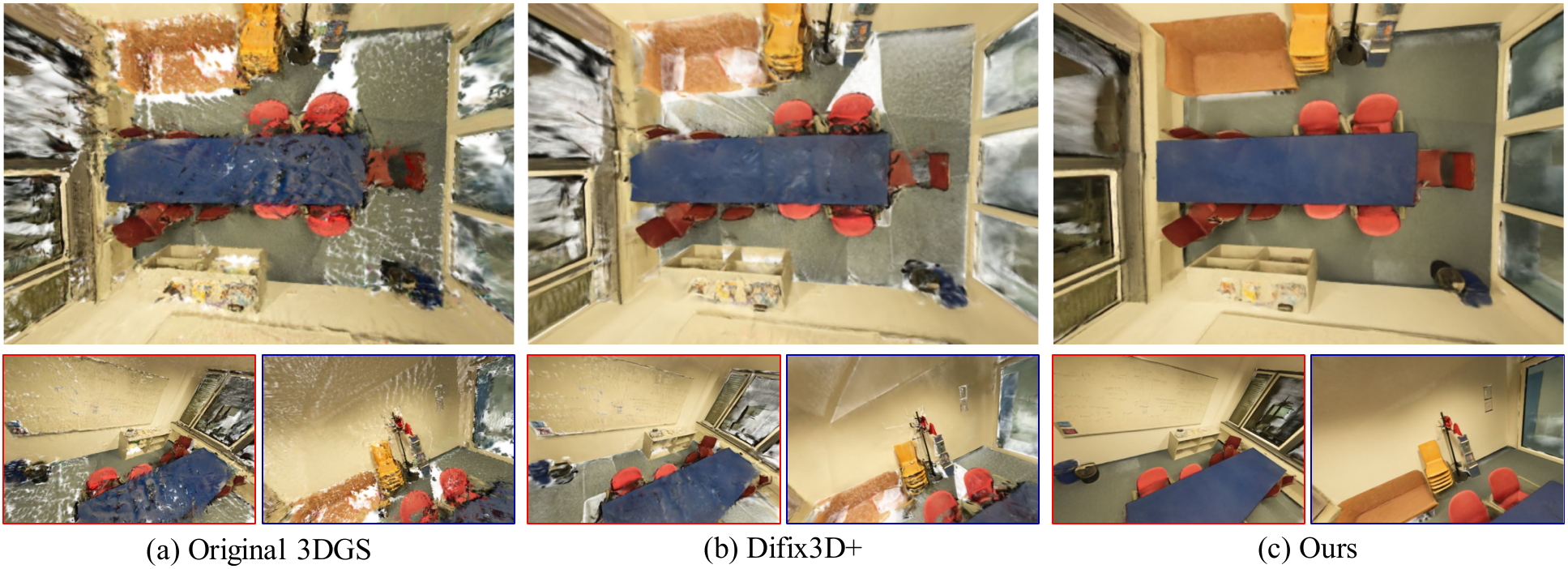}
  \caption{A room reconstructed by the original 3DGS \cite{kerbl20233d}, Difix3D+~\cite{wu2025difix3d+}, and our method, using only 8 images. Compared to the state-of-the-art method Difix3D+, our approach effectively recovers high-quality scenes without missing regions, blurring, or severe artifacts.
  }
  \label{fig:teaser}
\end{teaserfigure}

\begin{abstract}

We introduce S2C-3D, a novel sparse-view 3D reconstruction framework for high-fidelity and complete scene reconstruction from as few as six to eight images. Our framework features three components: a specialized diffusion model for scene-specific image restoration, a training-free view-consistency conditioned sampling process in the diffusion model for refined Gaussian optimization, and a camera trajectory planning scheme to ensure comprehensive scene coverage. The specialized diffusion model is developed by finetuning a pretrained architecture on the input views and their corresponding degraded counterparts. The adaptation to the scene distribution allows the model to repair Gaussian renderings while effectively eliminating domain gaps. Meanwhile, the trajectory planning scheme optimizes scene coverage by connecting each newly sampled camera to its two nearest neighbors. By iteratively constructing paths and retaining only those that significantly enhance visibility, the scheme establishes a trajectory that covers the entire scene. To address multi-view conflicts, the view-consistency conditioned sampling process quantifies the consistency between neighboring repaired images. This information is injected as a condition into the sampling process of the frozen diffusion model, facilitating the generation of view-consistent images without additional training. Consequently, our approach produces high-fidelity 3D Gaussians that are robust to artifacts. Experimental results demonstrate that S2C-3D outperforms state-of-the-art methods, constructing high-quality scenes that are free from missing regions, blurring, or other artifacts with very sparse inputs.
The source code and data are available at \textcolor{magenta}{ \href{https://gapszju.github.io/S2C-3D}{https://gapszju.github.io/S2C-3D}}.
%
%

%

\end{abstract}

%
%

\ccsdesc[500]{Computing methodologies~Reconstruction; Rendering; Point-based models}

%
%


\maketitle
{
\renewcommand{\thefootnote}{\fnsymbol{footnote}}

\footnotetext[2]{This is the author's version of The Work. It is posted here for your personal use. Not for redistribution.}
}
\section{Introduction}
\label{sec:intro}

Significant progress has been achieved in high-fidelity 3D scene reconstruction, particularly following the emergence of Neural Radiance Fields (NeRF)~\cite{mildenhall2021nerf} and 3D Gaussian Splatting (3DGS)~\cite{kerbl20233d}. While these methods enable the reconstruction of high-quality scenes from dozens of multi-view images, the data acquisition process remains time-consuming and labor-intensive. Furthermore, capturing such datasets is often challenging for novice users who lack expertise in view selection and coverage. Simplifying the capture process is therefore essential to broaden the application of these techniques in fields such as VR/AR, entertainment production, and autonomous robotics.

Some work has focused on 3D reconstruction from sparse-view images to streamline the acquisition process~\cite{li2024dngaussian,wu2025difix3d+,zhong2025taming,wu2025genfusion}.These approaches typically initialize low-quality 3D Gaussians from limited input views and subsequently enhance the images rendered from virtual cameras. These repaired images then serve as supervisory signals to refine the 3D Gaussian attributes. By iteratively performing image restoration and Gaussian refinement, these methods aim to produce high-quality scene representations. 
Early methods rely on different hand-crafted priors to restore Gaussian-rendered images, but are limited to reconstructing a localized region of a scene~\cite{hedman2021baking,yu2021plenoctrees,niemeyer2022regnerf,li2024dngaussian}.

More recently, using a pretrained diffusion model to iteratively improve the Gaussians has become a popular pipeline, and has achieved the state-of-the-art results~\cite{wu2025genfusion,wu2025difix3d+,zhong2025taming}.
However, when applying such a pipeline to the task of sparse-view 3D scene reconstruction, we observe that existing methods still struggle to generate high-quality results due to three key challenges that have been overlooked by prior approaches.
First, existing methods ignore the view inconsistency problem in diffusion-prior repaired images. While this may be acceptable for small unseen regions, it results in severe blurring and visual artifacts for large unseen regions (see the first row in Fig.~\ref{fig:supple_results2}).
Second, previous approaches for scene reconstruction~\cite{wu2025difix3d+,wu2025genfusion,zhong2025taming} use pretrained diffusion models without finetuning, leading to unsatisfying results when a large domain gap exists between the diffusion prior and the target scene (see the middle example in Fig.~\ref{fig:teaser}).
Third, existing frameworks employ simple camera interpolation between the input views~\cite{wu2025genfusion,wu2025difix3d+,zhong2025taming,li2024dngaussian}, where the generated virtual cameras only cover limited regions around these input viewpoints and fail to provide sufficient coverage of the entire 3D scene, resulting in incomplete reconstructions (see the left example in Fig.~\ref{fig:cameras}).

In this paper, we aim to reconstruct high-fidelity 3D scenes from \emph{extremely sparse inputs}, i.e., only $6$ to $8$ images. 
To tackle these challenges, we create a specialized diffusion model by finetuning a pretrained diffusion model~\cite{wu2025difix3d+} on the input and corresponding degraded images, adapting the model to the target scene so it can repair scene renderings without domain gap.
Then we design a simple but effective camera trajectory planning scheme. For each newly sampled virtual camera, we connect it to its two nearest neighboring cameras to build a candidate path, and measure the scene coverage of the path. By iteratively retaining only the camera paths that provide sufficient improvement in scene coverage, we obtain a camera trajectory that covers the whole scene.
In the key stage of Gaussian refinement, we freeze the diffusion model, measure the view consistency among neighboring repaired images, and inject it as a condition into the sampling process of the frozen diffusion model during Gaussian optimization. It progressively guides the diffusion model to generate more view-consistent images in a training-free manner, finally obtaining high-fidelity 3D Gaussians.

Experiments on various indoor scene datasets demonstrate that our method can reconstruct high-fidelity 3D scenes using only six to eight image captures.
Compared to state-of-the-art methods, our approach effectively recovers high-quality scenes without noticeable artifacts.

In summary, the main contributions of our work include:
\begin{itemize}
\item A novel framework for reconstructing high-fidelity 3D scenes using only 6-8 image captures.
\item A specialized diffusion model for scene image repair without domain gap, a view-consistency conditioned sampling process in the diffusion model for Gaussian refinement, and a camera trajectory planning scheme
that generates virtual cameras covering the entire 3D scene.
\item Superior performance compared with state-of-the-art methods on various indoor scene datasets in terms of both reconstruction quality and completeness.
\end{itemize}

\begin{figure*}[!ht] \centering
	\includegraphics[width=1\linewidth]{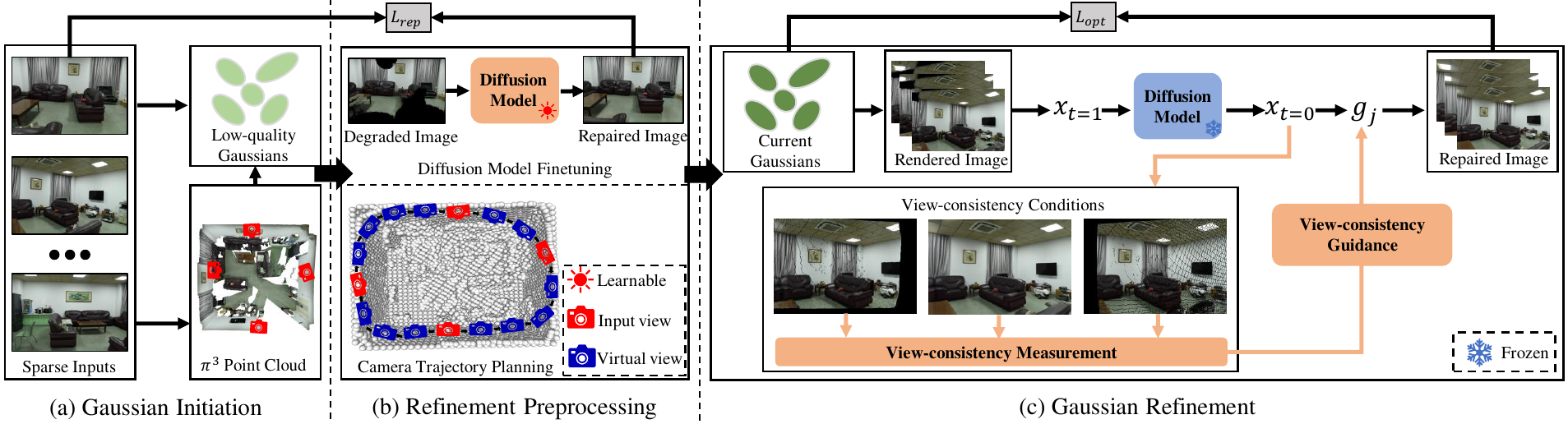}
	\caption{
 Overview of our method.
 (a) We input a sparse set of (e.g., 4) unposed images into a feed-forward visual geometry reconstruction model $\pi^{3}$ \cite{wang2025pi} to estimate camera poses and reconstruct a point cloud, which are then used to obtain an initial set of low-quality 3D Gaussians.
 (b) We create a specialized diffusion model by finetuning a pretrained diffusion model on the input and corresponding degraded images.
 %
Besides, we design a camera trajectory planning scheme to obtain a camera trajectory that covers the whole scene.
(c) We repair the Gaussian-rendered images at the planned camera trajectory, and use
the repaired images to optimize Gaussians for Gaussian refinement.
As the repaired images still have conflicts across different views, which cannot be directly used to generate high-quality Gaussians, we propose a training-free view-consistency conditioned sampling process in the diffusion model for Gaussian refinement.
 }
	\label{fig:oveview}
\end{figure*}
\section{Related Work}
\label{sec:related_work}
The success of Neural Radiance Fields (NeRF) \cite{mildenhall2021nerf} and 3D Gaussian Splatting (3DGS) \cite{kerbl20233d} has inspired a great amount of follow-up research for 3D scene reconstrcution \cite{barron2022mip,muller2022instant,ye2025gaussian,papantonakis2024reducing,huang20242d,kulhanek2024wildgaussians,peng2024rtg,sabour2025spotlesssplats,matsuki2024gaussian,fu2024colmap,yu2024mip,roessle2023ganerf,ren2024scube,peng2025gaussian}.
\cite{chen2024pgsr,wu2024recent,gao2022nerf} offer comprehensive surveys of recent progress in this area.
In this paper, we focus only on works related to sparse-view 3D reconstruction.

\subsection{Prior-based 3D scene reconstruction}
Early approaches typically leverage priors, such as smoothness losses \cite{hedman2021baking,yu2021plenoctrees}, sparsity priors \cite{niemeyer2022regnerf,yang2023freenerf,zhang2024fregs}, and depth-normal guidance \cite{li2024dngaussian,han2025sparserecon,zheng2025nexusgs}, to repair the Gaussian-rendered images at virtual cameras.
For example, Yu \textit{et al.} \cite{yu2021plenoctrees} introduce a smoothness prior on the 3D Gaussian density field to encourage spatially coherent geometry, reducing noise and artifacts during sparse-view optimization.
Zhang \textit{et al.} \cite{zhang2024fregs} introduce a sparsity prior on the 3D Gaussian density to suppress unsupported Gaussians and prevent overfitting to sparse views. By encouraging a compact representation, the method reduces spurious geometry in unseen regions and stabilizes the optimization process under sparse-view settings.
Li \textit{et al.} \cite{li2024dngaussian} leverage global–local depth and normal priors to supervise the optimization of 3D Gaussians in unseen regions, enabling the recovery of accurate scene geometry and detailed appearance from sparse inputs.
While promising results are demonstrated, these methods can only deal with small regions.

\subsection{Diffusion model based 3D scene reconstruction} 
Inspired by the powerful generative ability of general diffusion models, recent methods leverage general diffusion models retrained on large-scale real-world scene datasets to repair the Gaussian-rendered images, including image diffusion models \cite{wu2025difix3d+} and video diffusion models \cite{wu2025genfusion,wu2025difix3d+,liu2024reconx,bao2025free360}.
However, these methods overlook the fact that the repaired scene images still have conflicts across different views, leading to blurring and apparent artifacts on large unseen regions.
Notably, FreeDoM~\cite{yu2023freedom} is designed to ensure consistency between the input text/image and the output image. VD-3DGS \cite{zhong2025taming} only enforces consistency inside each repaired image, i.e., the consistency between the repaired and observed regions. It ignores the consistency across repaired images, leading to unsatisfying results on large unseen regions. In contrast, we measure the consistency across multi-view images, and use the multi-view consistency to guide the sampling process.

Besides, most methods~\cite{wu2025difix3d+,wu2025genfusion,zhong2025taming} use pretrained diffusion models without finetuning, leading to unsatisfying results when large domain gap exists between the diffusion prior and target scene.
GaussianObject \cite{yang2024gaussianobject} 
is the only one performing finetuning, but it is designed for object-level reconstruction. Its finetuning strategy (i.e., adding noises on Gaussians) cannot simulate the large unseen regions of scenes.

Beyond sparse-view 3D scene reconstruction methods, several single-image-to-3D approaches use diffusion models to complete Gaussian-rendered images \cite{yu2025wonderworld,ni2025wonderturbo,zhang2025scene}, but they still suffer from the mentioned three challenges, which restrict them to reconstructing only small scene regions from a single input view.

\section{Method}
\label{Method}
The overview of our method is illustrated in Fig.~\ref{fig:oveview}, which consists of three stages.
Given a sparse set of (e.g., 4) unposed images as input, we feed them into the feed-forward visual geometry reconstruction model $\pi^{3}$~\cite{wang2025pi} to estimate camera poses and reconstruct a point cloud $P$, which are then used to obtain an initial set of low-quality 3D Gaussians $G_0$ (Sec.~\ref{sub:3.1}).
Next we create a specialized diffusion model by finetuning a pretrained diffusion model \cite{wu2025difix3d+} on the input and corresponding
degraded images, so that the general diffusion model is adapted to the target scene (Sec.~\ref{sub:3.2}).
%
Besides, we design a camera trajectory planning scheme to obtain a camera trajectory $\tau_{best}$ that covers the whole scene (Sec.~\ref{sub:3.3}).
%
As the repaired images still have conflicts across different views, which cannot be directly used to generate high-quality Gaussians, we propose a training-free view-consistency conditioned sampling process in the diffusion model for Gaussian refinement (Sec.~\ref{sub:3.4}). 
%
%
%
We next provide a brief overview of 3DGS and diffusion models (Sec.~\ref{sub:3.0}), after which we describe each component of our framework in detail.

\subsection{Preliminary}
\label{sub:3.0}
\subsubsection{3D Gaussian Splatting}
3DGS~\cite{kerbl20233d} explicitly models a 3D scene with a collection of anisotropic Gaussians. Each primitive is parameterized by a tuple $\{\boldsymbol{\mu}, \mathbf{s}, \mathbf{q}, \boldsymbol{\alpha}, \mathbf{c}\}$, comprising the position $\boldsymbol{\mu} \in \mathbb{R}^{3}$, scaling factors $\mathbf{s} \in \mathbb{R}^{3}$, rotation quaternion $\mathbf{q} \in \mathbb{R}^{4}$, opacity $\boldsymbol{\alpha} \in \mathbb{R}$, and view-dependent color coefficients $\mathbf{c}$. 
The spatial influence of the $i$-th Gaussian is defined as:
\begin{equation}
\label{g0v}
G_0(\mathbf{v})=\exp\left(-\frac{1}{2}(\mathbf{v}-\boldsymbol{\mu}_i)^{\top}\mathbf{\Sigma}_i^{-1}(\mathbf{v}-\boldsymbol{\mu}_i)\right)
\end{equation}
where the covariance matrix $\mathbf{\Sigma}_i$ is derived from the scaling and rotation parameters $\mathbf{s}_i$ and $\mathbf{q}_i$. 
For rendering, 3D Gaussians are projected onto the image plane to form 2D splats, denoted as $G_0'$, which are then composited via front-to-back $\boldsymbol{\alpha}$-blending. 
The final pixel color $C(\mathbf{v}_p)$ is formulated as:
\begin{equation}
\label{3dgs_refined}
    C(\mathbf{v}_p) = \sum_{i \in \mathcal{K}} \mathbf{c}_i \sigma_i \prod_{j=1}^{i-1}(1-\sigma_j), \qquad \sigma_i = \boldsymbol{\alpha}_i G_0'(\mathbf{v}_p),
\end{equation}
where $\mathcal{K}$ denotes the ordered set of Gaussians overlapping pixel $\mathbf{v}_p$, sorted by depth.

\subsubsection{Diffusion model} 
Diffusion models first corrupt clean samples by progressively adding Gaussian noise and subsequently adopt a denoising function to reconstruct the original data by the sampling process \cite{ho2020denoising,rombach2022high,saharia2022palette,zhang2023sine,ho2022video,blattmann2023stable}. 
The underlying principle of standard diffusion models involves employing a noise estimator $\mathbf{\epsilon}_{\theta}$, typically a U-Net \cite{ronneberger2015u}, to predict the noise injected during the iterative $T$-step sampling process \cite{yu2023freedom,ho2020denoising} as follows:
\begin{equation}\label{denoising}
\mathbf{x}_{t-1}=\left(1+\beta_{t} / 2\right) \mathbf{x}_{t}+\beta_{t} \nabla_{\mathbf{x}_{t}} \log p\left(\mathbf{x}_{t}\right)+\sqrt{\beta_{t}} \mathbf{z},
\end{equation}
where $\nabla_{\mathbf{x}_t} \log p(\mathbf{x}_t)$ represents the estimated score function derived from $\boldsymbol{\epsilon}_{\theta}(\mathbf{x}_t, t)$. In this formulation, $\beta_t$ denotes predefined variance parameters and $\mathbf{z} \sim \mathcal{N}(\mathbf{0}, \mathbf{I})$ accounts for the stochastic noise component.
%
%
%
Given condition $\mathcal{Q}$, SDE \cite{song2020score} modifies the original score function $\nabla_{\mathbf{x}_{t}} \log p\left(\mathbf{x}_{t}\right)$ in Eq.~\ref{denoising} to a conditional score function $\nabla_{\mathbf{x}_{t}} \log p\left(\mathbf{x}_{t} \mid \mathcal{Q}\right)$ to guide the sampling process, which can be expanded using the Bayesian rule as:
\begin{equation}
\label{bayesian}
\nabla_{\mathbf{x}_{t}} \log p\left(\mathbf{x}_{t} \mid \mathcal{Q}\right)=\nabla_{\mathbf{x}_{t}} \log p\left(\mathbf{x}_{t}\right)+\nabla_{\mathbf{x}_{t}} \log p\left(\mathcal{Q} \mid \mathbf{x}_{t}\right),
\end{equation}
where $\nabla_{\mathbf{x}_{t}} \log p\left(\mathcal{Q} \mid \mathbf{x}_{t}\right)$ acts as a correction gradient so that the generated results are compatible with the given condition $\mathcal{Q}$.
If $\mathcal{Q}$ is defined as the view-consistency condition, $\nabla_{\mathbf{x}_{t}} \log p\left(\mathcal{Q} \mid \mathbf{x}_{t}\right)$ can be used to guide the diffusion model sampling toward generating view-consistent images.

\subsection{Gaussian Initiation}\label{sub:3.1}
Given unposed inputs, the initial low-quality Gaussians $G_0$ typically rely on point clouds and camera poses estimated by Structure-from-Motion (SfM) \cite{schonberger2016structure,schonberger2016pixelwise,garcia2017hierarchical}. 
However, traditional SfM yields inaccurate camera pose estimates and incomplete or noisy point cloud for sparse inputs.
Alternatively, we adopt the $\pi^{3}$ model \cite{wang2025pi}, a recent feed-forward visual geometry reconstruction framework for estimating affine-invariant camera poses and scale-invariant point clouds from sparse inputs, as illustrated in Fig.~\ref{fig:oveview}~(a).
$\pi^{3}$ gives a higher-quality initial point cloud and more reliable camera poses. Based on its prediction, we obtain the initial Gaussians $G_0$ by minimizing:
\begin{equation}
\label{opt_loss}
\mathcal{L}_{opt} = \left\|\mathbf{x}_{gt}-\mathbf{x}_{r}(G_0)\right\|_{1}+ (1-SSIM(\mathbf{x}_{gt},\mathbf{x}_{r}(G_0)))
\end{equation}
where $\mathbf{x}_{gt}$ and $\mathbf{x}_{r}$ denote the sparse input images and the rendered images, respectively. 
$G_0$ is then used to finetune the general diffusion model and serves as the initialization for the Gaussian refinement. 
Meanwhile, the generated $\pi^{3}$ point cloud is utilized for subsequent camera trajectory planning.

\subsection{Diffusion Model Finetuning}
\label{sub:3.2}
%
%
%
Although the input views are sparse, they capture informative appearance characteristics of the scene. We therefore leverage these images to finetune a specialized diffusion model tailored to the scene context.
Specifically, we finetune the model using training pairs generated by degrading the available input views. We propose a masking technique for rendered images to represent occluded or invisible areas. This approach effectively creates a mapping between the degraded observations and the original scene content for scene-specific adaptation.
As in \cite{yang2024gaussianobject}, we add random noise to the attributes of $G_0$ and render noise-degraded images from the noisy Gaussians at all reference views.
%
Those noisy Gaussians alone cannot model occluded and invisible regions.
%
To this end, we introduce masks by randomly removing a subset of Gaussians to simulate degraded images with large occlusion for better finetuning.
Following \cite{sauer2024adversarial, wu2025difix3d+}, we finetune the model with a frozen VAE encoder and a LoRA-adapted decoder.
The overall loss function used to supervise the training process is defined as:
\begin{equation}
\label{loss_phase2}
\mathcal{L}_{rep} = \left\|R(\mathbf{x}_{deg})-\mathbf{x}_{gt}\right\|_{2}+\mathcal{L}_{gram}+\mathcal{L}_{LPIPS},
\end{equation}
where we adopt an $L_2$ loss and a perceptual LPIPS loss between the model prediction $R(\mathbf{x}_{deg})$ and the sparse input $\mathbf{x}_{gt}$, along with a Gram matrix loss $\mathcal{L}_{gram}$ \cite{wu2025difix3d+}.

\begin{algorithm}[t]
\caption{Camera Trajectory Planning.}
\label{alg:init_vif}
\begin{algorithmic}[1] 
\STATE \textbf{input}: incomplete poisson mesh, input cameras $\tau_{input} = {\{\tau_{j}\}}_{j=1}^{m}$, added threshold $\varphi$, maximum step $Q_{end}$, interpolation threshold $d_{\varphi}$
\STATE \textbf{output}: best trajectory $\tau_{best}$
\STATE Sample spheres $\{S_{i}\}_{i=1}^{N}$ on the poisson mesh and its OBB.
\STATE Sample points ${\{s_{i}^{k}\}}_{k=1}^{N'}$ on each sphere $S_{i}$.
\WHILE{$Q < Q_{end}$}
    \STATE $\tau_{k} \leftarrow$ Randomly sample a camera in OBB
    \STATE $\{\tau_{k_1},\tau_{k_2}\} \leftarrow$ Find two nearest cameras in $\tau_{input}$ via Eq.\ref{Dist}
    \STATE $m_1 \leftarrow \text{Int}(\text{Dist}(\tau_{k_1}, \tau_{k})/d_{\varphi})$, $m_2 \leftarrow \text{Int}(\text{Dist}(\tau_{k_2}, \tau_{k})/d_{\varphi}$)
    \STATE $\{\tau_{k_1=k-m_{1}}, \ldots, \tau_{k}, \ldots, \tau_{k_2=k+m_{2}}\}\leftarrow$ Trajectory by Eq. \ref{inter}
    \STATE $g_{w} \leftarrow \text{Gain}(\tau_{k_1}, \ldots, \tau_{k}, \ldots, \tau_{k_2}|\tau_{input})$
    \IF{$g_{w}>\varphi$}
        \STATE $\tau_{input} \leftarrow \tau_{input} \bigcup \{\tau_{k_1}, \ldots, \tau_{k}, \ldots, \tau_2\}$
        \STATE $Q \leftarrow 0$
    \ELSE
        \STATE $Q \leftarrow Q+1$
    \ENDIF
\ENDWHILE
\STATE $\tau_{best} \leftarrow $ Output a camera trajectory $\tau_{input}$ with sufficient coverage
\end{algorithmic}
\end{algorithm}

\begin{figure}\centering
	\includegraphics[width=1\linewidth]{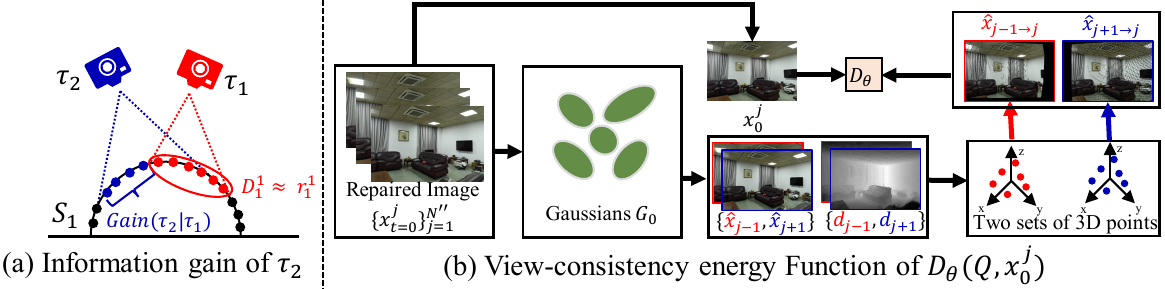}
	\caption{Detailed illustrations of the computation of the information gain (Sec.~\ref{sub:3.3}) and the view-consistency energy function (Sec.~\ref{sub:3.4}).
    }
\label{fig:details}
\end{figure}
\subsection{Camera Trajectory Planning}
\label{sub:3.3}
%

%
To ensure sufficient coverage of the full 3D scene, inspired by \cite{xie2018creating,roberts2016generating,zhang2021continuous,yang2018uncut}, we propose a camera trajectory planning scheme. We first define an information gain metric to quantify the contribution of each camera to scene coverage, followed by a description of how this metric is used for camera trajectory planning.
%
%

\subsubsection{Information gain} We first adopt Poisson surface reconstruction~\cite{kazhdan2006poisson} to convert the $\pi^{3}$ point cloud $P$ into a surface mesh.
As the $\pi^{3}$ point cloud $P$ is generated from sparse views, the reconstructed surface mesh may not be watertight. 
If we directly follow~\cite{kazhdan2006poisson,roberts2017submodular} and sample points only from the mesh surface, these incomplete regions cannot be represented. 
To solve it, we sample more points from the six faces of an oriented bounding box (OBB) that tightly encloses the coarse mesh to cover missing regions.
%
%
We convert all of the $N$ surface points to spheres $\{S_i\}_{i=1}^{N}$ to cover the scene (see Fig. \ref{fig:oveview} (b)).
%
%
%
%
After that, we compute the intersection area between these spheres ${\{S_{i}\}}_{i=1}^{N}$ and the view frustum of each camera $\tau_j$ to evaluate how well $\tau_j$ contributes to the overall scene coverage.
%
%
%
Concretely, the coverage of the local area around $S_i$ from camera $\tau_{j}$ is denoted as $D_i^j$,
which corresponds to the intersection between the viewing frustum of $\tau_{j}$ and the sphere $S_i$.
The effective area of $D_i^j$ depends on both the distance between $\tau_{j}$ and $S_i$, as well as the elevation angle of the camera relative to the surface.
To simplify the computation of the intersection area, we uniformly sample $N'$ surface points ${\{s_{i}^{k}\}}_{k=1}^{N'}$ on each sphere $S_i$, where each point $s_{i}^{k}$ covers a local sphere area.
As shown in Fig. \ref{fig:details} (a), the coverage of the local area by $D_i^j$ can be represented as the collection of sample points $r_i^j$ on the sphere $S_i$ that fall within the view frustum of $\tau_{j}$:
\begin{equation}
\label{disk}
D_i^j \approx r_i^j = {\{s_{i}^{k} \mid s_{i}^{k} \in S_i \land s_{i}^{k} \in D_i^j\}}_{k=1}^{N'}
\end{equation}
if $\tau_{j}$ is defined as a candidate camera to be appended to the planned camera path, which consists of a sequence of cameras $\{\tau_1, \ldots, \tau_{j-1}\}$.
The contribution of $\tau_{j}$ to covering all 3D contents, also called \textit{information gain} \cite{kazhdan2006poisson}, is defined as the amount of added scene information. 
It denotes the number of new sample points on each local area $D_i^j$ that can be observed by $\tau_j$ beyond the observations from the previous cameras $\{\tau_1, \ldots, \tau_{j-1}\}$:
\begin{equation}
\label{gain}
\text{Gain}(\tau_j|\tau_{j-1},...,\tau_{1}) = \sum_{i} \left| r_i^j \setminus \bigcup_{l=1}^{j-1} r_i^l \right|
\end{equation}

\subsubsection{Camera trajectory planning}
The detailed camera trajectory planning scheme is outlined in Alg.~\ref{alg:init_vif}. We first record the information gain of the input cameras $\tau_{input} = {\{\tau_{j}\}}_{j=1}^{m}$, and randomly sample a new camera $\tau_{k}$ in OBB.
Then, we compute the pose distance between the new camera $\tau_{k}$ and each camera in the set $\tau_{input}$, which can be formulated as:
\begin{equation}
\label{Dist}
\text{Dist}(\tau_1, \tau_2) = w_t \cdot \| t_1 - t_2 \|_2 + w_r \cdot \arccos \left( 2 \left( \vert q_1 \cdot q_2 \vert \right)^2 - 1 \right),
\end{equation}
where $\tau_1$ and $\tau_2$ denote the poses of two cameras. $w_t$ and $w_r$ represent the weights for the translation and rotation distances, respectively.
$t_1$ and $t_2$ denote the translation vectors, while $q_1$ and $q_2$ represent the rotation vectors.
According to Eq.~\ref{Dist}, we obtain the two cameras $\{\tau_{k_1},\tau_{k_2}\}$ in $\tau_{input}$ with the smallest pose distances to $\tau_{k}$, and interpolate a trajectory of length $m_1 + m_2 + 1$ between $\tau_{k}$ and $\{\tau_{k_1},\tau_{k_2}\}$ as follows:
\begin{equation}
\label{inter}
\{\tau_{k_1}, \ldots, \tau_{k}, \ldots, \tau_{k_2}\} = \text{inter}(\tau_{k_1}, \tau_{k}) \cup \text{inter} (\tau_{k}, \tau_{k_2}).
\end{equation}
We extend Eq. \ref{gain} to estimate the information gain $g_{w}$ of the interpolated trajectory $\{\tau_{k_1}, \ldots, \tau_{k}, \ldots, \tau_{k_2}\}$ and use it to determine whether the entire interpolated trajectory should be included in the final trajectory.
If the information gain $g_w$ exceeds a predefined threshold $\varphi$, all cameras in the interpolated trajectory are considered to provide sufficient new information and are therefore incorporated into the set $\tau_{input}$. 
This process is iteratively repeated until no additional cameras can be found that contribute new information to the scene.
Thus, $\tau_1$ and $\tau_2$ in the updated $\tau_{input}$ are not only input poses, but are also already sampled and adopted virtual camera poses. The interpolation among these cameras randomly sampled inside the scene OBB can cover more regions than existing methods.
Finally, we obtain the optimal trajectory $\tau_{best}$, which allows a small number of cameras to fully reconstruct the 3D scene, while ensuring that each planned camera receives complementary guidance information from two neighboring repaired images, thereby propagating view consistency across all planned cameras (see Sec. \ref{sub:3.4}).

\begin{algorithm}[t]
\caption{View-consistency Conditioned Sampling Process.}
\label{alg:refinement}
\begin{algorithmic}[1] 
\STATE \textbf{input}: initial low-quality Gaussians $G_0$, trajectory $\tau_{best}$, refine steps $r_{t}$
\STATE \textbf{output}: output 3DGS $G^*$
\FOR{$i=r_{t},...,1$}
    \STATE $\{{\mathbf{x}_{t=1}^{j}}\}_{j=1}^{N''} \leftarrow$ Render $N''$ images by $G_0$ via $\tau_{best}$
    \FOR{$j=N'',...,1$}
     \STATE $\mathbf{x}_{0}^{j} \leftarrow \left(1+\beta_{1} / 2\right) \mathbf{x}_{1}^{j}+\beta_{1} \nabla_{\mathbf{x}_{1}^{j}} \log p\left(\mathbf{x}_{1}^{j}\right)+\sqrt{\beta_{1}} \mathbf{z}$
    \ENDFOR
    \STATE $G_0\leftarrow$ Optimize 3DGS using $\{\mathbf{x}_{0}^{j}\}_{j=1}^{N''}$
    \IF{i=1}
    \STATE break
    \ENDIF
    \FOR{$j=N'',...,1$}
    \STATE$\boldsymbol{g}_{j} \leftarrow \nabla_{\mathbf{x}_{t=1}^{j}} \mathcal{D}_{\boldsymbol{\theta}}\left(\mathcal{Q}, \mathbf{x}_{t=0}^{j}\right)$
    \STATE$\mathbf{x}_{0}^{j} \leftarrow \mathbf{x}_{0}^{j}-\rho_{j} \boldsymbol{g}_{j}$, where $\rho_j$ is a predefined scale factor
    \ENDFOR
    \STATE $G_0\leftarrow$ Optimize 3DGS using $\{\mathbf{x}_{0}^{j}\}_{j=1}^{N''}$
\ENDFOR
\STATE $G^* \leftarrow$ Output 3DGS $G_0$
\end{algorithmic}
\end{algorithm}

\subsection{Gaussian Refinement}
\label{sub:3.4}
%
Previous work, e.g., Difix3D+ and GenFusion, directly conditions the diffusion model on multiple views. However, as mentioned above, these methods cannot guarantee the view consistency cross multi-view images.
Hence, we propose a training-free view-consistency conditioned sampling process in the diffusion model for Gaussian refinement.
During Gaussian refinement, we freeze the diffusion model, measure the view
consistency among neighboring repaired images, and inject it as a condition into the sampling process of the frozen diffusion model. 
Doing so progressively guides the diffusion
model to generate more view-consistent images, finally obtaining high-fidelity 3D Gaussians $G^*$.

\subsubsection{Diffusion guidance} As discussed in Sec.~\ref{sub:3.0}, by defining the view-consistency condition as $\mathcal{Q}$, the corresponding conditional score term $\nabla_{\mathbf{x}_{t}} \log p\left(\mathcal{Q} \mid \mathbf{x}_{t}\right)$ in Eq.~\ref{bayesian} can be used to enforce view consistency among the repaired images.
%
%
Inspired by \cite{yu2023freedom}, we adopt a training-free manner to inject the view-consistency condition into the sampling process in the diffusion model.
Especially, the term $\nabla_{\mathbf{x}_{t}} \log p\left(\mathcal{Q} \mid \mathbf{x}_{t}\right)$ can further be modeled as a consistency-guided energy function \cite{lecun2006tutorial,zhao2022egsde,yu2023freedom}:
\begin{equation}
\label{energy}
 p\left(\mathcal{Q} \mid \mathbf{x}_{t}\right)=\frac{\exp \left\{-\lambda \mathcal{E}\left(\mathcal{Q},\mathbf{x}_{t}\right)\right\}}{Z},
\end{equation}
where $\lambda$ denotes a positive temperature coefficient and $Z>0$ is a normalization constant. $\mathcal{E}(\mathcal{Q}, \mathbf{x}_{t})$ represents an energy function that quantifies how well the current sample $\mathbf{x}_{t}$ aligns with the target of view-consistency. The energy function between $\mathcal{Q}$ and $\mathbf{x}_{t}$ can be further approximated by the time-dependent energy function between $\mathcal{Q}$ and $\mathbf{x}_{0}$ \cite{yu2023freedom} as follows:
\begin{equation}
\label{guide}
\mathcal{E}(\mathcal{Q}, \mathbf{x}_{t}) \approx \mathcal{D}_{\boldsymbol{\theta}}\left(\mathcal{Q}, \mathbf{x}_{0 \mid t}\right).
\end{equation}
With Eq. \ref{energy} and Eq. \ref{guide}, it can be written as:
\begin{equation}
\label{guide_final}
\nabla_{\mathbf{x}_{t}} \log p\left(\mathcal{Q} \mid \mathbf{x}_{t}\right) \propto-\nabla_{\mathbf{x}_{t}} \mathcal{D}_{\boldsymbol{\theta}}\left(\mathcal{Q}, \mathbf{x}_{0}\right).
\end{equation}
Notably,  we adopt the single-step (i.e., $T=1$) diffusion model Difix3D+ \cite{wu2025difix3d+}, which takes the images rendered from the initial low-quality Gaussians $G_0$ as inputs, 
denoted as $\{\mathbf{x}_{t=1}^{j}\}_{j=1}^{N''}=\{\mathbf{x}_{t=1}^{1},...,\mathbf{x}_{t=1}^{N''}\}$,
to generate the repaired images $\{\mathbf{x}_{t=0}^{j}\}_{j=1}^{N''}$.
Thus, the problem can be reformulated as defining a suitable energy function $\mathcal{D}_{\boldsymbol{\theta}}\left(\mathcal{Q}, \mathbf{x}_{0}\right)$ that effectively measures the consistency among the multi-view repaired images $\{\mathbf{x}_{0}^{j}\}_{j=1}^{N''}$.
%

\subsubsection{Energy formulation} In order to build such a consistency measurement function between $\{\mathbf{x}_{0}^{j}\}_{j=1}^{N''}$, we establish pixel-wise correspondences across views and then measure the discrepancies between corresponding pixels. 
Reliable correspondences can be derived from depth maps and associated camera poses~\cite{schonberger2016structure,mildenhall2021nerf,toft2020single}. We leverage these geometric cues to project pixels into 3D points and unproject them across views to construct accurate cross-view correspondences.
As shown in Fig. \ref{fig:details} (b), 
we first use $\{\mathbf{x}_{0}^{j}\}_{j=1}^{N''}$ to optimize 3DGS $G_0$ and then render corresponding RGB images $\{\hat{\mathbf{x}}_{j}\}_{j=1}^{N''}$ and depth images $\{d_{j}\}_{j=1}^{N''}$. 
For each image $\mathbf{x}_{0}^{j}$, we unproject its two neighboring rendered images $\hat{\mathbf{x}}_{j-1}$ and $\hat{\mathbf{x}}_{j+1}$ using their corresponding depth maps $d_{j-1}$ and $d_{j+1}$, together with the camera poses, to obtain two sets of 3D RGB points. These points are then projected onto the image plane of $\mathbf{x}_{0}^{j}$ to generate two point-rendered images $\hat{\mathbf{x}}_{j-1\rightarrow j}$ and $\hat{\mathbf{x}}_{j+1\rightarrow j}$. 
%
%
$\hat{\mathbf{x}}_{j-1 \rightarrow j}$ and $\hat{\mathbf{x}}_{j+1 \rightarrow j}$ provide complementary multi-view information for $\mathbf{x}_{0}^{j}$. Therefore, we typically select these two neighboring point-rendered images for view-consistency measurement. 
%
%
The final consistency energy function is formulated as the pixel-wise L1 loss between each original repaired image $\mathbf{x}_{0}^{j}$ and its corresponding neighboring point-rendered images $\{\hat{\mathbf{x}}_{j-1 \rightarrow j}, \hat{\mathbf{x}}_{j+1 \rightarrow j}\}$:
\begin{equation}
\label{depth}
\mathcal{D}_{\boldsymbol{\theta}}\left(\mathcal{Q}, \mathbf{x}_{0}^{j}\right)
=
\left\| \mathbf{M}_{j-1} \odot \left( \mathbf{x}_{0}^{j}- \hat{\mathbf{x}}_{j-1\rightarrow j} \right) \right\|_{1}
+
\left\| \mathbf{M}_{j+1} \odot \left( \mathbf{x}_{0}^{j}- \hat{\mathbf{x}}_{j+1\rightarrow j} \right) \right\|_{1},
\end{equation}
where $\mathbf{M}_{j-1}$ and $\mathbf{M}_{j+1}$ denote the validity masks corresponding to the point-rendered images $\hat{\mathbf{x}}_{j-1 \rightarrow j}$ and $\hat{\mathbf{x}}_{j+1 \rightarrow j}$, respectively, and $\odot$ represents element-wise multiplication. 
As shown in Alg.~\ref{alg:refinement}, this view-consistency energy function can inject view-consistency conditions into the sampling process during Gaussian optimization, to progressively produce more view-consistent images (See Fig.~\ref{fig:view-consistency}) in a training-free manner.

\definecolor{best}{RGB}{255, 199, 199}   
\definecolor{second}{RGB}{255, 230, 199} 
\definecolor{third}{RGB}{255, 255, 199}  
\begin{table*}[!t]
\centering
\footnotesize
\caption{
Quantitative comparisons of sparse-view 3D reconstruction methods on the ScanNet++, Replica, and S2C-Scene datasets. 
Our approach demonstrates strong performance across diverse domains and outperforms all compared methods. 
We highlight the results as follows: \colorbox{best}{best}, \colorbox{second}{second best}, and \colorbox{third}{third best}.
}
\begin{tabular}{c|c|ccc|ccc|ccc}
\toprule
\multirow{3}{*}{Dataset} & \multirow{3}{*}{Method}& \multicolumn{3}{c|}{6-view} & \multicolumn{3}{c|}{7-view} & \multicolumn{3}{c}{8-view} \\
\cmidrule(lr){3-5} \cmidrule(lr){6-8} \cmidrule(lr){9-11}
& & PSNR $\uparrow$ & SSIM $\uparrow$ &LPIPS $\downarrow$ &   PSNR $\uparrow$ & SSIM $\uparrow$ &LPIPS $\downarrow$ &  PSNR $\uparrow$ & SSIM $\uparrow$&LPIPS $\downarrow$  \\
\midrule
\multirow{5}{*}{\rotatebox{90}{ScanNet++}}
& 3DGS \cite{kerbl20233d}& 13.063 &0.496 & 0.465 & 13.911 & 0.517 & 0.452& 14.956 & 0.554 & 0.432 \\
& DNGaussian \cite{li2024dngaussian}&12.834  &0.420  &0.571  & 13.338 & 0.436 & 0.560 & 13.728 & 0.446 & 0.551 \\
& Difix3D+ \cite{wu2025difix3d+}& 13.321 & 0.549 &  0.405 & 14.228 & 0.575 & 0.386 & 15.300 & 0.613 & 0.361 \\
& Genfusion \cite{wu2025genfusion}& \colorbox{second}{16.605} & \colorbox{third}{0.626} & \colorbox{second}{0.371} & \colorbox{third}{17.139} & \colorbox{third}{0.661} & \colorbox{second}{0.355} & \colorbox{third}{17.993} & \colorbox{third}{0.683} & \colorbox{second}{0.331} \\
& VD-3DGS~\cite{zhong2025taming} & \colorbox{third}{16.536} & \colorbox{second}{0.644} & \colorbox{third}{0.395} & \colorbox{second}{17.488} & \colorbox{second}{0.667} & \colorbox{third}{0.368} & \colorbox{second}{18.291} & \colorbox{second}{0.688} & \colorbox{third}{0.346} \\
 &Ours & \colorbox{best}{17.779} & \colorbox{best}{0.683} & \colorbox{best}{0.344}& \colorbox{best}{17.990} & \colorbox{best}{0.690} & \colorbox{best}{0.333}  & \colorbox{best}{18.777} & \colorbox{best}{0.706} & \colorbox{best}{0.315} \\
\midrule
\multirow{3}{*}{Dataset} & \multirow{3}{*}{Method}& \multicolumn{3}{c|}{4-view} & \multicolumn{3}{c|}{5-view} & \multicolumn{3}{c}{6-view} \\
\cmidrule(lr){3-5} \cmidrule(lr){6-8} \cmidrule(lr){9-11}
& & PSNR $\uparrow$ & SSIM $\uparrow$ &LPIPS $\downarrow$ &   PSNR $\uparrow$ & SSIM $\uparrow$ &LPIPS $\downarrow$ &  PSNR $\uparrow$ & SSIM $\uparrow$&LPIPS $\downarrow$  \\ \midrule
\multirow{5}{*}{\rotatebox{90}{Replica}}
& 3DGS \cite{kerbl20233d}& 13.577 & 0.487 & 0.416 & 14.673 & 0.545 & 0.379 & 16.117 & 0.576 & 0.370 \\
& DNGaussian \cite{li2024dngaussian}& 14.264 & 0.460 & 0.517 & 15.010 & 0.476 & 0.517 & 15.945 & 0.511 & ~0.504\\
& Difix3D+ \cite{wu2025difix3d+}& 13.498 & 0.521 & 0.371 & 14.745 & 0.576 & 0.329 & 16.297& 0.617 & 0.312 \\
& Genfusion \cite{wu2025genfusion}& \colorbox{second}{17.148} & \colorbox{second}{0.641} & \colorbox{second}{0.353} & \colorbox{second}{18.764}& \colorbox{second}{0.685} & \colorbox{second}{0.305}& \colorbox{second}{19.855}& \colorbox{second}{0.701} & \colorbox{second}{0.290} \\
& VD-3DGS~\cite{zhong2025taming} & \colorbox{third}{17.017} &\colorbox{third}{0.618}  &\colorbox{third}{0.387} & \colorbox{third}{18.437} & \colorbox{third}{0.675}& \colorbox{third}{0.340} & \colorbox{third}{19.592} & \colorbox{third}{0.699} & \colorbox{third}{0.314}\\
 &Ours & \colorbox{best}{18.445}& \colorbox{best}{0.685} & \colorbox{best}{0.329} & \colorbox{best}{19.917} & \colorbox{best}{0.714} & \colorbox{best}{0.290}& \colorbox{best}{20.360} & \colorbox{best}{0.721} & \colorbox{best}{0.281}\\
\midrule
\multirow{5}{*}{\rotatebox{90}{S2C-Scene}}
& 3DGS \cite{kerbl20233d}& 13.131 & 0.448 & 0.511 & 14.015 & 0.473 & 0.501 & 14.264 & 0.482 & 0.496 \\
& DNGaussian \cite{li2024dngaussian}& 13.033 & 0.428 & 0.586& 13.314 & 0.442& 0.577 & 13.469 & 0.444 & 0.572 \\
& Difix3D+ \cite{wu2025difix3d+}& 13.302 & 0.493 & 0.460 & 14.326& 0.533 & 0.438 & 14.571 & 0.544& 0.429 \\
& Genfusion \cite{wu2025genfusion}& \colorbox{second}{16.029} & \colorbox{second}{0.614} & \colorbox{second}{0.410} & \colorbox{second}{16.774} & \colorbox{second}{0.642} & \colorbox{second}{0.391} & \colorbox{second}{17.188} & \colorbox{second}{0.648}& \colorbox{second}{0.382}\\
& VD-3DGS~\cite{zhong2025taming} & \colorbox{third}{15.902} & \colorbox{third}{0.613} & \colorbox{third}{0.429} & \colorbox{third}{15.932} & \colorbox{third}{0.614} & \colorbox{third}{0.416}& \colorbox{third}{16.341} & \colorbox{third}{0.623} & \colorbox{third}{0.410} \\
 &Ours & \colorbox{best}{16.474} & \colorbox{best}{0.634} & \colorbox{best}{0.401} & \colorbox{best}{17.226} & \colorbox{best}{0.648} & \colorbox{best}{0.380} & \colorbox{best}{17.647} & \colorbox{best}{0.658} & \colorbox{best}{0.368} \\
\bottomrule
\end{tabular}
\label{tab:compare}
\end{table*}

\section{Experiments}
\label{Experiments}

\subsection{Experimental Setup}

\subsubsection{Baselines and metrics} We compare S2C-3D with recent sparse-view 3D scene reconstruction methods, including DNGaussian~\cite{li2024dngaussian}, Difix3D+~\cite{wu2025difix3d+}, GenFusion~\cite{wu2025genfusion}, and VD-3DGS~\cite{zhong2025taming}. 
For quantitative evaluation of novel view synthesis, we report PSNR, SSIM~\cite{wang2004image}, and LPIPS~\cite{zhang2018unreasonable}.
%



\subsubsection{Implementation details}
For diffusion model finetuning, we adopt Difix3D+~\cite{wu2025difix3d+} as the baseline and set the number of training steps to 3000. We use the AdamW optimizer with a constant learning rate of $2\times10^{-5}$ and a linear warmup of 500 steps.
All stages are trained on a RTX~4090 GPU with a 9950X3D CPU. 
The parameters are set as follows: the added threshold $\varphi = 0.1$, the maximum step $Q_{\text{end}} = 30$, the interpolation threshold $d_{\varphi} = 0.2$, the refinement steps $r_t = 4$, the consistency weight $\rho_j=1$ and the distance weights $w_t = 1$ and $w_r = 1$.
%
%

\subsubsection{Datasets} 
We conduct experiments on all 8 scenes from the synthetic Replica dataset \cite{straub2019replica}, 100 scenes from the real-world ScanNet++ dataset \cite{yeshwanth2023scannet++}, and 11 real-world scenes captured by ourselves, termed S2C-Scene.
The training-view settings of ScanNet++ are set to 6, 7 and 8 views. And the settings of Replica and S2C-Scene are set to 4, 5 and 6 views.

%
\subsection{Evaluation of Scene Reconstruction}
We qualitatively compare the rendering quality for novel view synthesis across all compared methods, as shown in Tab.~\ref{tab:compare}. 
We also evaluate the scene completeness on the ScanNet++ dataset in the supplementary materials.
The results show that our approach consistently achieves the best performance across all evaluation metrics, indicating superior reconstruction quality and scene completeness.

\begin{figure*}[ht]
  \centering
  \begin{minipage}{0.9\linewidth}
  \includegraphics[width = \linewidth]{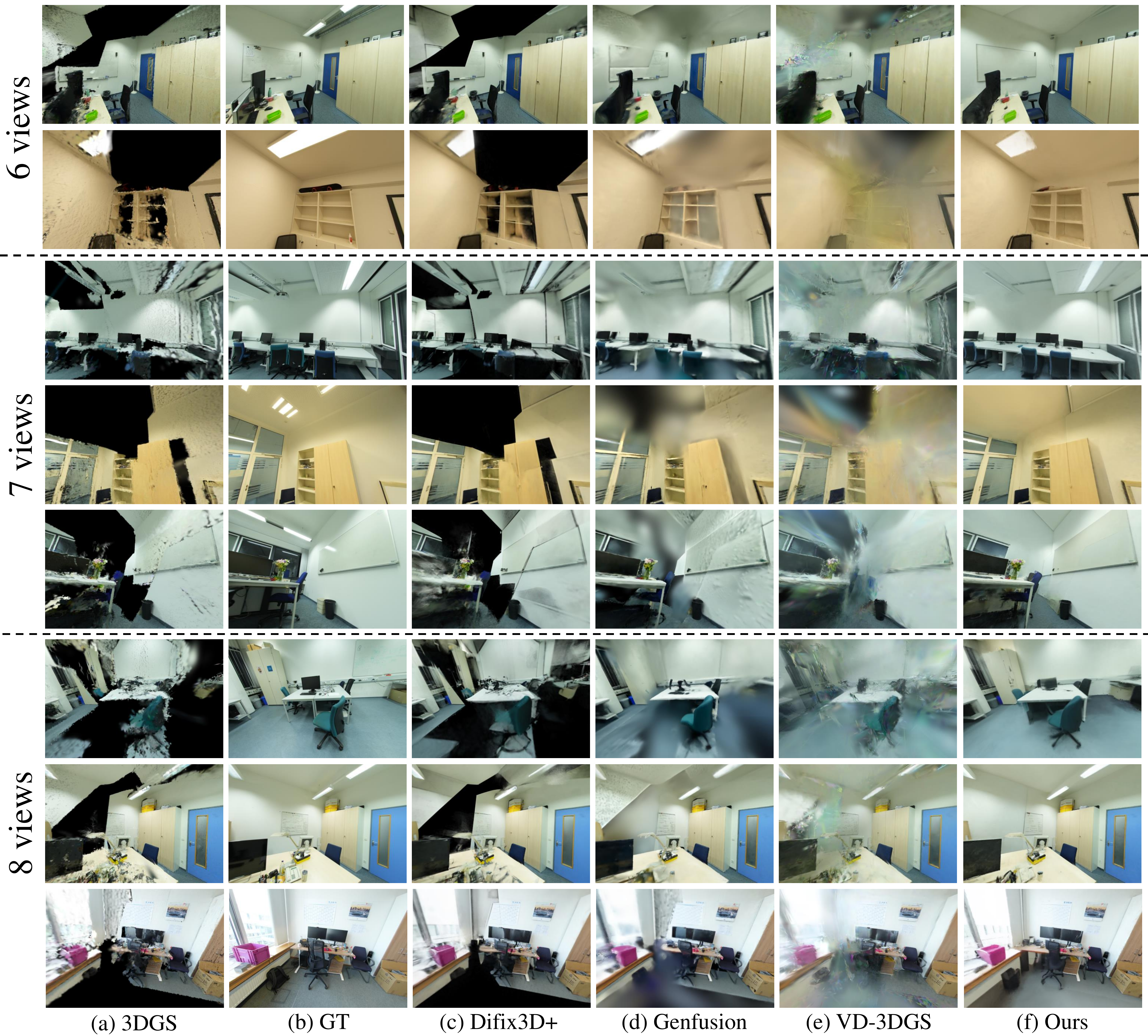}
  \caption{Qualitative comparisons of novel view synthesis on the ScanNet++ dataset.}
  \label{fig:supple_results}
  \end{minipage}
\begin{minipage}{0.9\linewidth}
  \includegraphics[width = \linewidth]{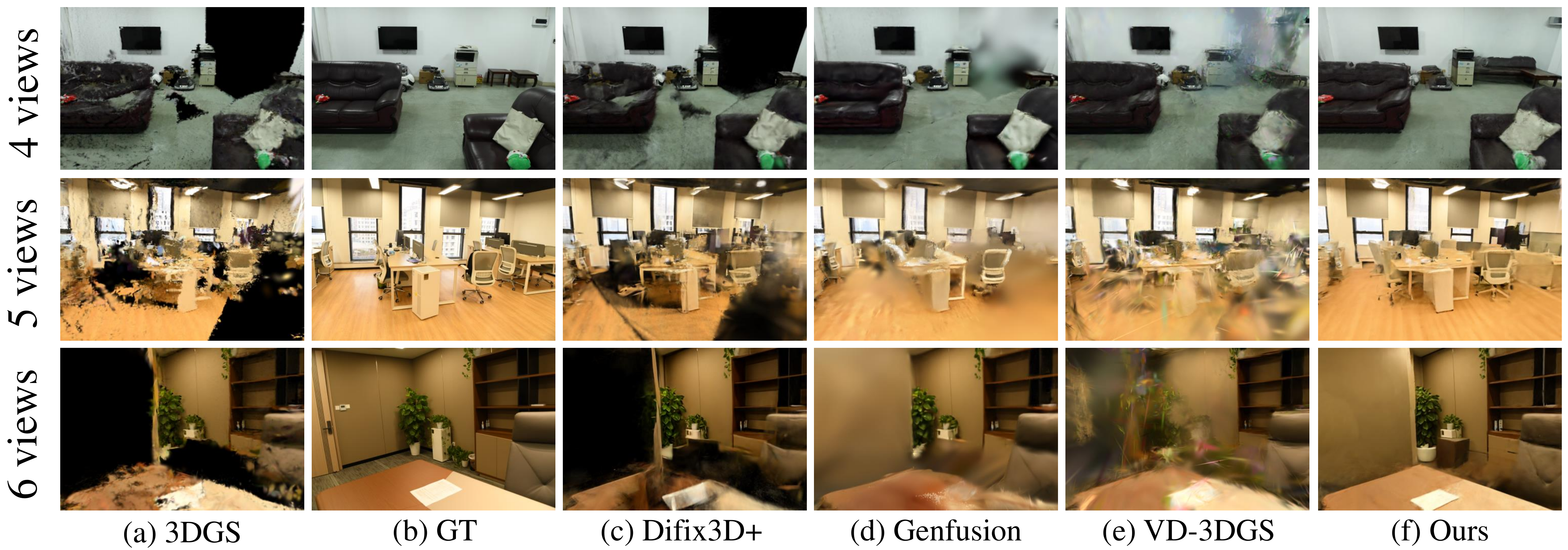}
  \caption{Qualitative comparisons of novel view synthesis on the S2C-Scene dataset.}
  \label{fig:supple_results2}
  \end{minipage}
\end{figure*}

Figs. \ref{fig:supple_results} and \ref{fig:supple_results2} show novel view synthesis results on the ScanNet++ and S2C-Scene dataset across different views.
We observe that all compared methods struggle to reconstruct high-fidelity scenes from highly sparse images. 
Especially, the results of Difix3D+, GenFusion, and VD-3DGS still contain large incomplete regions (e.g., the ceiling in the fourth row of Fig.~\ref{fig:supple_results}).
This is because the large domain gap between the training data and the target scene leads to a substantial repair performance drop (see examples of incomplete repaired images produced by Difix3D+ in Fig.~\ref{fig:supp_repaired}).
Besides, their simple interpolation between the input views cannot generate virtual cameras to cover these occluded and invisible regions (see Fig. \ref{fig:cameras}). 
By leveraging a specialized diffusion model for scene image repair without domain gap and a camera trajectory planning scheme that generates virtual cameras covering the entire 3D scene, our method enables complete 3D scene reconstruction.
Moreover, Genfusion tends to generate blurred Gaussians (e.g., the chair in the third row of Fig.~\ref{fig:supple_results}), and VD-3DGS often produces low-quality Gaussians with extensive artifacts (e.g., the table and background in the third row of Fig.~\ref{fig:supple_results2}). Both issues can also be observed in the 3D Gaussians generated by Difix3D+, such as the table in the third row of Fig.~\ref{fig:supple_results2}.
This is because they overlook the fact that the repaired scene images still have conflicts across different views. While our method proposes a view-consistency conditioned sampling process to progressively guide the diffusion model to generate more view-consistent images, finally obtaining high-fidelity 3D Gaussians.

\begin{table}[!t]
\centering
\footnotesize
\caption{Ablation study. w/o and w denote without and with models, respectively.
\textit{Fine.}, \textit{Traj.}, and \textit{Cond.} denote the proposed diffusion model finetuning, the camera trajectory planning scheme, and the view-consistency conditioned sampling process, respectively.
}
\setlength{\tabcolsep}{1mm}{
\begin{tabular}{c|ccc|ccc}
\toprule
\multirow{2}{*}{Model} &\multirow{2}{*}{\textit{Fine.}} &\multirow{2}{*}{\textit{Traj.}} &\multirow{2}{*}{\textit{Cond.}} & \multicolumn{3}{c}{Replica}  \\
&&&& PSNR$\uparrow$ & SSIM$\uparrow$ & LPIPS$\downarrow$\\
\midrule
Full Model       & \checkmark & \checkmark &      \checkmark &   20.360        & 0.721        &0.281     \\ \hline
w/o Noise Finetuning     &\checkmark  & \checkmark & \checkmark     & 19.815&	0.714&	0.295\\
w/o Mask Finetuning      &\checkmark  & \checkmark & \checkmark     & 18.522         & 0.653      &  0.336      \\
w Difix3D+        & $\times$  & \checkmark & \checkmark     & 18.160&	0.640	&0.349 \\ \hline
w/o Camera Trajectory Planning       & \checkmark & $\times$ &      \checkmark &   18.253    &    0.645    &   0.342    \\ \hline
w/o View-consistency Conditions       & \checkmark & \checkmark &  $\times$     &18.129         &  0.639      &  0.353    \\
\bottomrule 
\end{tabular}
}
\label{tab:ab}
\end{table}
\subsection{Ablation Study}
In this section, we isolate our algorithmic choices and evaluate their effects on the 6-view setting of the Replica dataset.

\subsubsection{The effect of diffusion model finetuning} 
To validate the effect of the specialized diffusion model, we conduct experiments on three models, e.g., without noise training pairs for finetuning, without mask training pairs for finetuning, and directly using the original Difix3D+ model without finetuning, as shown in the third to fifth rows of Tab.~\ref{tab:ab}.
It can be seen that the proposed specialized diffusion model achieves the best performance.
As shown in Fig.~\ref{fig:supp_repaired}, removing noise training pairs leads to degraded repair quality, while removing mask training pairs, or directly using Difix3D+, fails to repair large visible and occluded regions in the Gaussian-rendered images.
In conclusion, the proposed specialized diffusion model can adapt the model to the target scene so it can repair Gaussian renderings without domain gap to obtain high-quality Gaussians (see the first row of Fig.~\ref{fig:abalation}).


\subsubsection{The effect of camera trajectory planning} 
As shown in the sixth row of Tab.~\ref{tab:ab}, removing the proposed camera trajectory planning scheme and using simple interpolation between input views to generate virtual cameras for Gaussian refinement leads to a significant performance drop.
As shown in Fig.~\ref{fig:cameras}, the virtual cameras generated by our camera trajectory planning scheme cover more 3D content than those obtained by simple camera interpolation, resulting in more complete scene reconstruction (see the second and third rows of Fig.~\ref{fig:abalation}).

\subsubsection{The effect of view-consistency conditioned sampling process} 
\begin{figure*}[htbp]
  \centering
  \begin{minipage}{\linewidth}
    \centering
    \includegraphics[width=1\linewidth]{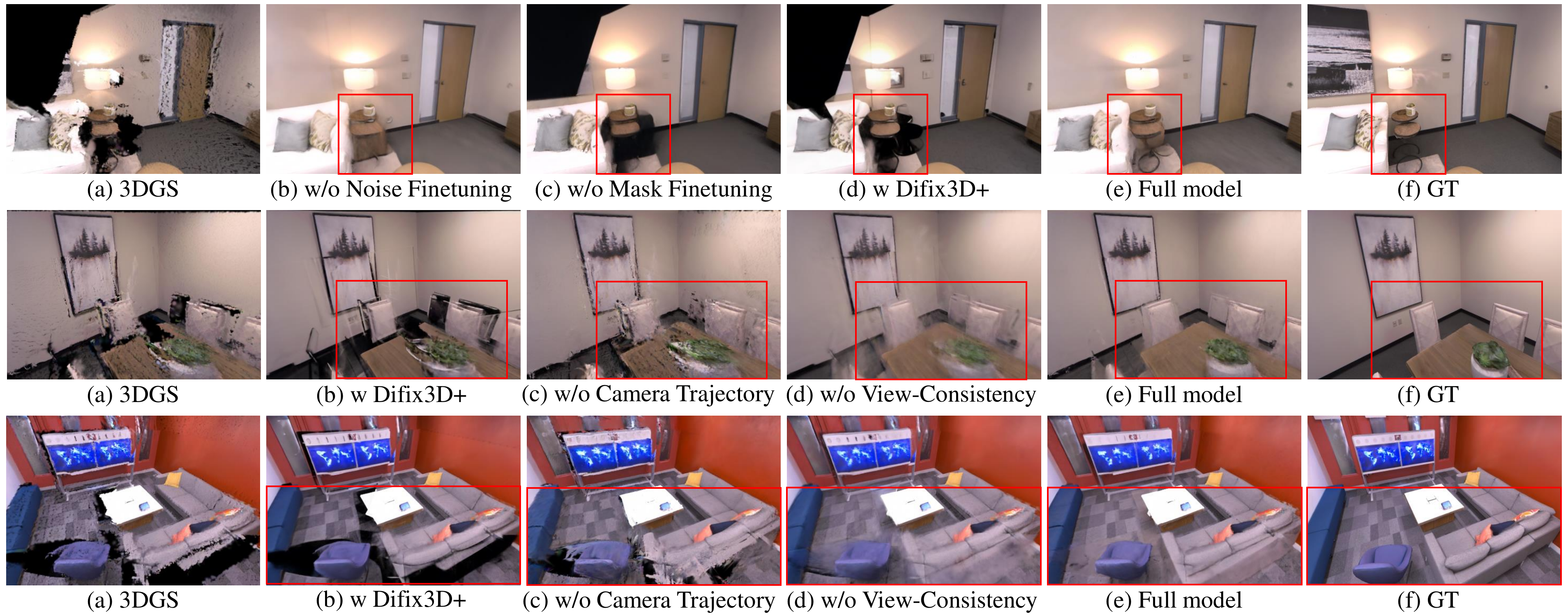}
    \caption{Ablation study of reconstructed scenes by different models.}
    \label{fig:abalation}
  \end{minipage}
\begin{minipage}{\linewidth}
    \centering
    \includegraphics[width=1\linewidth]{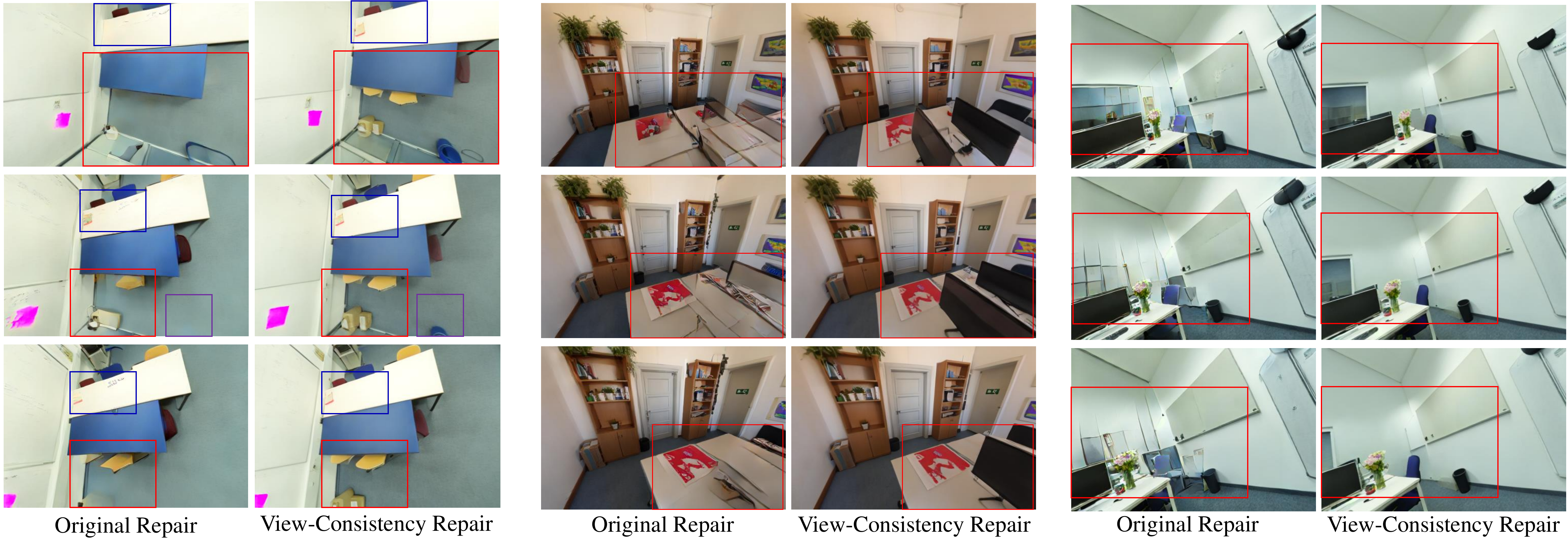}
    \caption{Qualitative comparison between images repaired by the finetuned diffusion model and those produced by the view-consistency conditioned diffusion process.}
    \label{fig:view-consistency}
  \end{minipage}

  \begin{minipage}{0.6\linewidth}
    \centering
    \includegraphics[width=\linewidth]{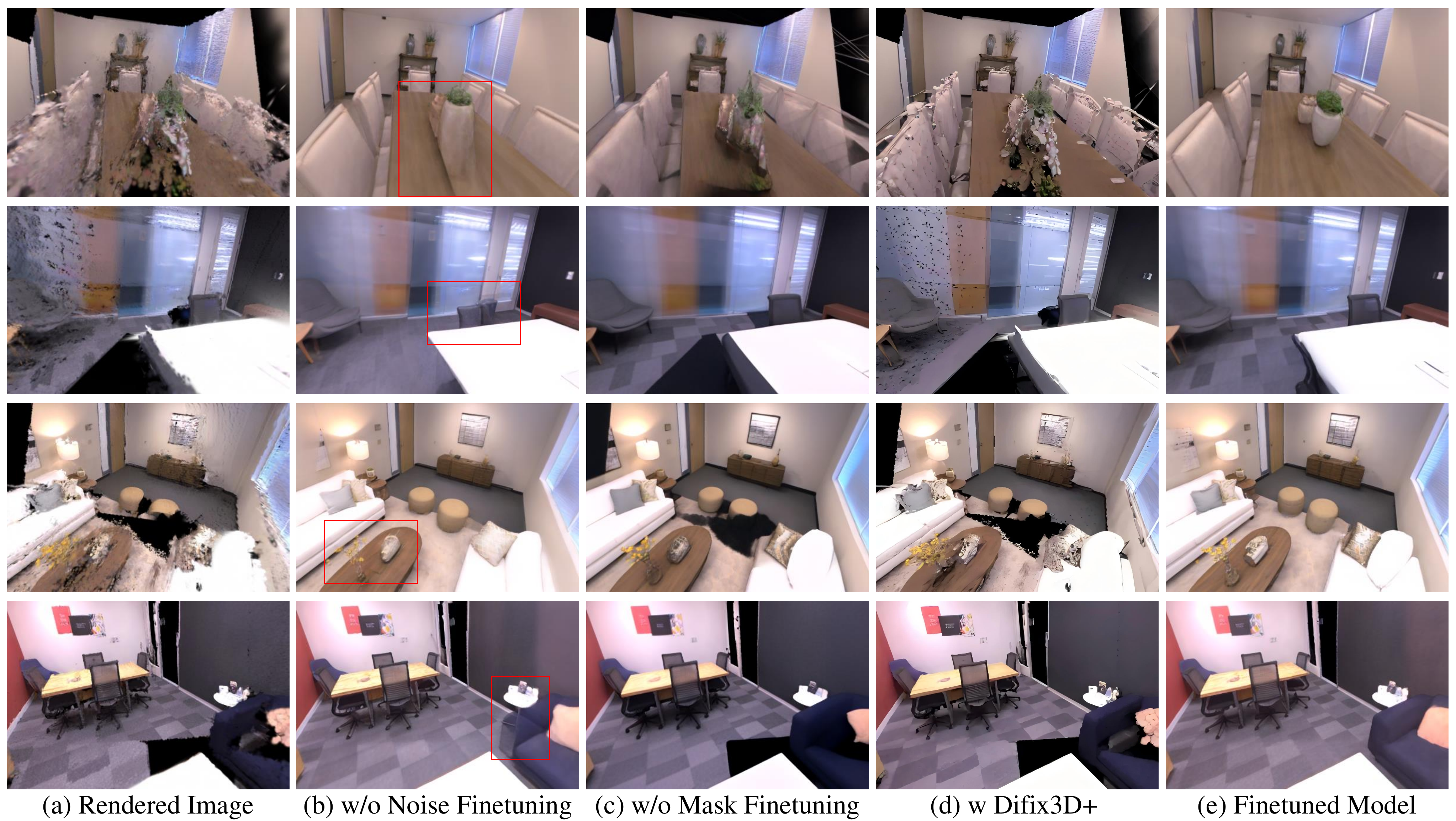}
    \caption{Qualitative comparison between images repaired by different diffusion models.}
    \label{fig:supp_repaired}
  \end{minipage}
  \begin{minipage}{0.39\linewidth}
    \centering
    \includegraphics[width=\linewidth]{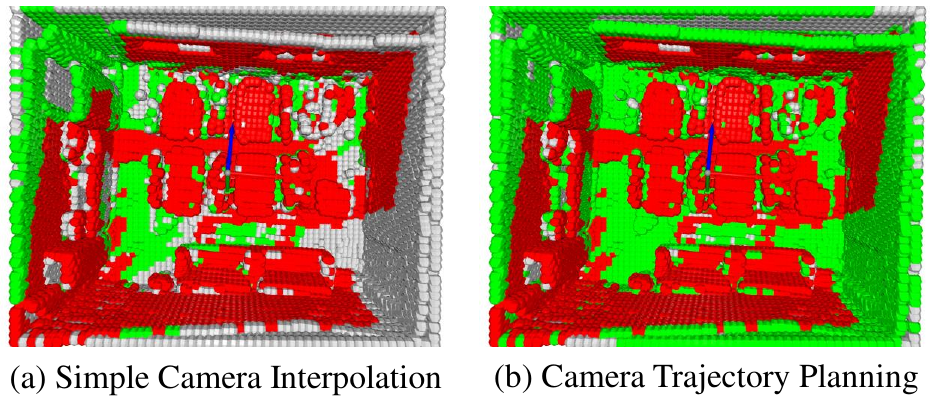}
    \caption{Comparison of 3D scene coverage achieved by virtual cameras generated via simple interpolation between input views and those obtained using our camera trajectory planning scheme.
    \textcolor{red}{Red} spheres are visible in the sparse input views, while \textcolor{green}{green} spheres indicate newly visible regions introduced by the virtual cameras generated by simple camera interpolation and our method.
   \textcolor{gray}{Gray} spheres represent regions that are invisible to all cameras.
    }
    \label{fig:cameras}
  \end{minipage}
  
\end{figure*}

As shown in the seventh row of Tab.~\ref{tab:ab}, removing the proposed view-consistency conditioned sampling process and directly using the repaired images from the finetuned diffusion model for Gaussian refinement results in a significant performance drop.
As shown in Fig.~\ref{fig:view-consistency}, the original repaired images still have conflicts across different views, which cannot be directly used to generate high-quality Gaussians.
While the proposed training-free view-consistency conditioned sampling process progressively guides the diffusion model to generate more view-consistent images (see  Fig.~\ref{fig:view-consistency}), finally obtaining high-quality Gaussians without blurring or extensive artifacts (see the second and third rows of Fig.~\ref{fig:abalation}).

\section{Discussion and Conclusion}
\label{Conclusion}
%
Our method focuses on indoor scenes with closed areas. For outdoor scenes with open areas, similar to existing methods, our method only reconstructs the regions among the input views.
Besides, the sampled cameras can occasionally fall into undesirable positions, e.g., inside objects, and under tables looking upward at the bottom surfaces. The corresponding regions have lower quality due to the limited information from the given views. But since these areas are less important and don't affect the main areas, it is acceptable to leave the undesired cameras there. Our method will fail on concave rooms, producing unexpected regions outside the groundtruth walls. 
The proposed camera trajectory planning scheme is a greedy algorithm. We will explore a better solution in future.

In conclusion, We propose a novel sparse-view 3D reconstruction framework, termed S2C-3D, to enable high-fidelity and complete 3D scene reconstruction from as few as 6-8 image captures. Our framework features a specialized diffusion model for scene image repair without domain gap, a training-free view-consistency conditioned sampling process in the diffusion model for Gaussian refinement, and a camera trajectory planning scheme that generates virtual cameras covering the entire 3D scene.

\begin{acks}
The authors would like to thank the reviewers for their insightful
comments. 
This work is supported by NSF China (No. U23A20311 \& 62322209 \& 62421003), and the XPLORER PRIZE.
\end{acks}
%

\bibliographystyle{ACM-Reference-Format}
\bibliography{main.bbl}


\begin{thebibliography}{63}


\ifx \showCODEN    \undefined \def \showCODEN     #1{\unskip}     \fi
\ifx \showISBNx    \undefined \def \showISBNx     #1{\unskip}     \fi
\ifx \showISBNxiii \undefined \def \showISBNxiii  #1{\unskip}     \fi
\ifx \showISSN     \undefined \def \showISSN      #1{\unskip}     \fi
\ifx \showLCCN     \undefined \def \showLCCN      #1{\unskip}     \fi
\ifx \shownote     \undefined \def \shownote      #1{#1}          \fi
\ifx \showarticletitle \undefined \def \showarticletitle #1{#1}   \fi
\ifx \showURL      \undefined \def \showURL       {\relax}        \fi
\providecommand\bibfield[2]{#2}
\providecommand\bibinfo[2]{#2}
\providecommand\natexlab[1]{#1}
\providecommand\showeprint[2][]{arXiv:#2}

\bibitem[Bao et~al\mbox{.}(2025)]%
        {bao2025free360}
\bibfield{author}{\bibinfo{person}{Chong Bao}, \bibinfo{person}{Xiyu Zhang}, \bibinfo{person}{Zehao Yu}, \bibinfo{person}{Jiale Shi}, \bibinfo{person}{Guofeng Zhang}, \bibinfo{person}{Songyou Peng}, {and} \bibinfo{person}{Zhaopeng Cui}.} \bibinfo{year}{2025}\natexlab{}.
\newblock \showarticletitle{Free360: Layered Gaussian Splatting for Unbounded 360-Degree View Synthesis from Extremely Sparse and Unposed Views}. In \bibinfo{booktitle}{\emph{2025 IEEE/CVF Conference on Computer Vision and Pattern Recognition (CVPR)}}. \bibinfo{pages}{16377--16387}.
\newblock
\href{https://doi.org/10.1109/CVPR52734.2025.01527}{doi:\nolinkurl{10.1109/CVPR52734.2025.01527}}


\bibitem[Barron et~al\mbox{.}(2022)]%
        {barron2022mip}
\bibfield{author}{\bibinfo{person}{Jonathan~T. Barron}, \bibinfo{person}{Ben Mildenhall}, \bibinfo{person}{Dor Verbin}, \bibinfo{person}{Pratul~P. Srinivasan}, {and} \bibinfo{person}{Peter Hedman}.} \bibinfo{year}{2022}\natexlab{}.
\newblock \showarticletitle{Mip-NeRF 360: Unbounded Anti-Aliased Neural Radiance Fields}. In \bibinfo{booktitle}{\emph{2022 IEEE/CVF Conference on Computer Vision and Pattern Recognition (CVPR)}}. \bibinfo{pages}{5460--5469}.
\newblock
\href{https://doi.org/10.1109/CVPR52688.2022.00539}{doi:\nolinkurl{10.1109/CVPR52688.2022.00539}}


\bibitem[Blattmann et~al\mbox{.}(2023)]%
        {blattmann2023stable}
\bibfield{author}{\bibinfo{person}{Andreas Blattmann}, \bibinfo{person}{Tim Dockhorn}, \bibinfo{person}{Sumith Kulal}, \bibinfo{person}{Daniel Mendelevitch}, \bibinfo{person}{Maciej Kilian}, \bibinfo{person}{Dominik Lorenz}, \bibinfo{person}{Yam Levi}, \bibinfo{person}{Zion English}, \bibinfo{person}{Vikram Voleti}, \bibinfo{person}{Adam Letts}, {et~al\mbox{.}}} \bibinfo{year}{2023}\natexlab{}.
\newblock \showarticletitle{Stable video diffusion: Scaling latent video diffusion models to large datasets}.
\newblock \bibinfo{journal}{\emph{arXiv preprint arXiv:2311.15127}} (\bibinfo{year}{2023}).
\newblock


\bibitem[Chen et~al\mbox{.}(2025)]%
        {chen2024pgsr}
\bibfield{author}{\bibinfo{person}{Danpeng Chen}, \bibinfo{person}{Hai Li}, \bibinfo{person}{Weicai Ye}, \bibinfo{person}{Yifan Wang}, \bibinfo{person}{Weijian Xie}, \bibinfo{person}{Shangjin Zhai}, \bibinfo{person}{Nan Wang}, \bibinfo{person}{Haomin Liu}, \bibinfo{person}{Hujun Bao}, {and} \bibinfo{person}{Guofeng Zhang}.} \bibinfo{year}{2025}\natexlab{}.
\newblock \showarticletitle{PGSR: Planar-Based Gaussian Splatting for Efficient and High-Fidelity Surface Reconstruction}.
\newblock \bibinfo{journal}{\emph{IEEE Transactions on Visualization and Computer Graphics}} \bibinfo{volume}{31}, \bibinfo{number}{9} (\bibinfo{year}{2025}), \bibinfo{pages}{6100--6111}.
\newblock
\href{https://doi.org/10.1109/TVCG.2024.3494046}{doi:\nolinkurl{10.1109/TVCG.2024.3494046}}


\bibitem[Fu et~al\mbox{.}(2024)]%
        {fu2024colmap}
\bibfield{author}{\bibinfo{person}{Yang Fu}, \bibinfo{person}{Xiaolong Wang}, \bibinfo{person}{Sifei Liu}, \bibinfo{person}{Amey Kulkarni}, \bibinfo{person}{Jan Kautz}, {and} \bibinfo{person}{Alexei~A. Efros}.} \bibinfo{year}{2024}\natexlab{}.
\newblock \showarticletitle{COLMAP-Free 3D Gaussian Splatting}. In \bibinfo{booktitle}{\emph{2024 IEEE/CVF Conference on Computer Vision and Pattern Recognition (CVPR)}}. \bibinfo{pages}{20796--20805}.
\newblock
\href{https://doi.org/10.1109/CVPR52733.2024.01965}{doi:\nolinkurl{10.1109/CVPR52733.2024.01965}}


\bibitem[Gao et~al\mbox{.}(2022)]%
        {gao2022nerf}
\bibfield{author}{\bibinfo{person}{Kyle Gao}, \bibinfo{person}{Yina Gao}, \bibinfo{person}{Hongjie He}, \bibinfo{person}{Dening Lu}, \bibinfo{person}{Linlin Xu}, {and} \bibinfo{person}{Jonathan Li}.} \bibinfo{year}{2022}\natexlab{}.
\newblock \showarticletitle{Nerf: Neural radiance field in 3d vision, a comprehensive review}.
\newblock \bibinfo{journal}{\emph{arXiv preprint arXiv:2210.00379}} (\bibinfo{year}{2022}).
\newblock


\bibitem[Garcia-Fidalgo and Ortiz(2017)]%
        {garcia2017hierarchical}
\bibfield{author}{\bibinfo{person}{Emilio Garcia-Fidalgo} {and} \bibinfo{person}{Alberto Ortiz}.} \bibinfo{year}{2017}\natexlab{}.
\newblock \showarticletitle{Hierarchical Place Recognition for Topological Mapping}.
\newblock \bibinfo{journal}{\emph{IEEE Transactions on Robotics}} \bibinfo{volume}{33}, \bibinfo{number}{5} (\bibinfo{year}{2017}), \bibinfo{pages}{1061--1074}.
\newblock
\href{https://doi.org/10.1109/TRO.2017.2704598}{doi:\nolinkurl{10.1109/TRO.2017.2704598}}


\bibitem[Han et~al\mbox{.}(2025)]%
        {han2025sparserecon}
\bibfield{author}{\bibinfo{person}{Liang Han}, \bibinfo{person}{Xu Zhang}, \bibinfo{person}{Haichuan Song}, \bibinfo{person}{Kanle Shi}, \bibinfo{person}{Yu-Shen Liu}, {and} \bibinfo{person}{Zhizhong Han}.} \bibinfo{year}{2025}\natexlab{}.
\newblock \showarticletitle{SparseRecon: Neural Implicit Surface Reconstruction from Sparse Views with Feature and Depth Consistencies}. In \bibinfo{booktitle}{\emph{Proceedings of the IEEE/CVF International Conference on Computer Vision (ICCV)}}. \bibinfo{pages}{28514--28524}.
\newblock


\bibitem[Hedman et~al\mbox{.}(2021)]%
        {hedman2021baking}
\bibfield{author}{\bibinfo{person}{Peter Hedman}, \bibinfo{person}{Pratul~P. Srinivasan}, \bibinfo{person}{Ben Mildenhall}, \bibinfo{person}{Jonathan~T. Barron}, {and} \bibinfo{person}{Paul Debevec}.} \bibinfo{year}{2021}\natexlab{}.
\newblock \showarticletitle{Baking Neural Radiance Fields for Real-Time View Synthesis}. In \bibinfo{booktitle}{\emph{2021 IEEE/CVF International Conference on Computer Vision (ICCV)}}. \bibinfo{pages}{5855--5864}.
\newblock
\href{https://doi.org/10.1109/ICCV48922.2021.00582}{doi:\nolinkurl{10.1109/ICCV48922.2021.00582}}


\bibitem[Ho et~al\mbox{.}(2020)]%
        {ho2020denoising}
\bibfield{author}{\bibinfo{person}{Jonathan Ho}, \bibinfo{person}{Ajay Jain}, {and} \bibinfo{person}{Pieter Abbeel}.} \bibinfo{year}{2020}\natexlab{}.
\newblock \showarticletitle{Denoising Diffusion Probabilistic Models}. In \bibinfo{booktitle}{\emph{Advances in Neural Information Processing Systems}}, \bibfield{editor}{\bibinfo{person}{H.~Larochelle}, \bibinfo{person}{M.~Ranzato}, \bibinfo{person}{R.~Hadsell}, \bibinfo{person}{M.F. Balcan}, {and} \bibinfo{person}{H.~Lin}} (Eds.), Vol.~\bibinfo{volume}{33}. \bibinfo{publisher}{Curran Associates, Inc.}, \bibinfo{pages}{6840--6851}.
\newblock
\urldef\tempurl%
\url{https://proceedings.neurips.cc/paper_files/paper/2020/file/4c5bcfec8584af0d967f1ab10179ca4b-Paper.pdf}
\showURL{%
\tempurl}


\bibitem[Ho et~al\mbox{.}(2022)]%
        {ho2022video}
\bibfield{author}{\bibinfo{person}{Jonathan Ho}, \bibinfo{person}{Tim Salimans}, \bibinfo{person}{Alexey Gritsenko}, \bibinfo{person}{William Chan}, \bibinfo{person}{Mohammad Norouzi}, {and} \bibinfo{person}{David~J Fleet}.} \bibinfo{year}{2022}\natexlab{}.
\newblock \showarticletitle{Video Diffusion Models}. In \bibinfo{booktitle}{\emph{Advances in Neural Information Processing Systems}}, \bibfield{editor}{\bibinfo{person}{S.~Koyejo}, \bibinfo{person}{S.~Mohamed}, \bibinfo{person}{A.~Agarwal}, \bibinfo{person}{D.~Belgrave}, \bibinfo{person}{K.~Cho}, {and} \bibinfo{person}{A.~Oh}} (Eds.), Vol.~\bibinfo{volume}{35}. \bibinfo{publisher}{Curran Associates, Inc.}, \bibinfo{pages}{8633--8646}.
\newblock
\urldef\tempurl%
\url{https://proceedings.neurips.cc/paper_files/paper/2022/file/39235c56aef13fb05a6adc95eb9d8d66-Paper-Conference.pdf}
\showURL{%
\tempurl}


\bibitem[Huang et~al\mbox{.}(2024)]%
        {huang20242d}
\bibfield{author}{\bibinfo{person}{Binbin Huang}, \bibinfo{person}{Zehao Yu}, \bibinfo{person}{Anpei Chen}, \bibinfo{person}{Andreas Geiger}, {and} \bibinfo{person}{Shenghua Gao}.} \bibinfo{year}{2024}\natexlab{}.
\newblock \showarticletitle{2D Gaussian Splatting for Geometrically Accurate Radiance Fields}. In \bibinfo{booktitle}{\emph{ACM SIGGRAPH 2024 Conference Papers}} \emph{(\bibinfo{series}{SIGGRAPH '24})}. \bibinfo{publisher}{Association for Computing Machinery}, \bibinfo{address}{New York, NY, USA}, Article \bibinfo{articleno}{32}, \bibinfo{numpages}{11}~pages.
\newblock
\showISBNx{9798400705250}
\href{https://doi.org/10.1145/3641519.3657428}{doi:\nolinkurl{10.1145/3641519.3657428}}


\bibitem[Kazhdan et~al\mbox{.}(2006)]%
        {kazhdan2006poisson}
\bibfield{author}{\bibinfo{person}{Michael Kazhdan}, \bibinfo{person}{Matthew Bolitho}, {and} \bibinfo{person}{Hugues Hoppe}.} \bibinfo{year}{2006}\natexlab{}.
\newblock \showarticletitle{Poisson surface reconstruction}. In \bibinfo{booktitle}{\emph{Proceedings of the Fourth Eurographics Symposium on Geometry Processing}} (Cagliari, Sardinia, Italy) \emph{(\bibinfo{series}{SGP '06})}. \bibinfo{publisher}{Eurographics Association}, \bibinfo{address}{Goslar, DEU}, \bibinfo{pages}{61–70}.
\newblock
\showISBNx{3905673363}


\bibitem[Kerbl et~al\mbox{.}(2023)]%
        {kerbl20233d}
\bibfield{author}{\bibinfo{person}{Bernhard Kerbl}, \bibinfo{person}{Georgios Kopanas}, \bibinfo{person}{Thomas Leimkuehler}, {and} \bibinfo{person}{George Drettakis}.} \bibinfo{year}{2023}\natexlab{}.
\newblock \showarticletitle{3D Gaussian Splatting for Real-Time Radiance Field Rendering}.
\newblock \bibinfo{journal}{\emph{ACM Trans. Graph.}} \bibinfo{volume}{42}, \bibinfo{number}{4}, Article \bibinfo{articleno}{139} (\bibinfo{date}{July} \bibinfo{year}{2023}), \bibinfo{numpages}{14}~pages.
\newblock
\showISSN{0730-0301}
\href{https://doi.org/10.1145/3592433}{doi:\nolinkurl{10.1145/3592433}}


\bibitem[Kulhanek et~al\mbox{.}(2024)]%
        {kulhanek2024wildgaussians}
\bibfield{author}{\bibinfo{person}{Jonas Kulhanek}, \bibinfo{person}{Songyou Peng}, \bibinfo{person}{Zuzana Kukelova}, \bibinfo{person}{Marc Pollefeys}, {and} \bibinfo{person}{Torsten Sattler}.} \bibinfo{year}{2024}\natexlab{}.
\newblock \showarticletitle{WildGaussians: 3D Gaussian Splatting In the Wild}. In \bibinfo{booktitle}{\emph{Advances in Neural Information Processing Systems}}, \bibfield{editor}{\bibinfo{person}{A.~Globerson}, \bibinfo{person}{L.~Mackey}, \bibinfo{person}{D.~Belgrave}, \bibinfo{person}{A.~Fan}, \bibinfo{person}{U.~Paquet}, \bibinfo{person}{J.~Tomczak}, {and} \bibinfo{person}{C.~Zhang}} (Eds.), Vol.~\bibinfo{volume}{37}. \bibinfo{publisher}{Curran Associates, Inc.}, \bibinfo{pages}{21271--21288}.
\newblock
\href{https://doi.org/10.52202/079017-0670}{doi:\nolinkurl{10.52202/079017-0670}}


\bibitem[LeCun et~al\mbox{.}(2006)]%
        {lecun2006tutorial}
\bibfield{author}{\bibinfo{person}{Yann LeCun}, \bibinfo{person}{Sumit Chopra}, \bibinfo{person}{Raia Hadsell}, \bibinfo{person}{M Ranzato}, \bibinfo{person}{Fujie Huang}, {et~al\mbox{.}}} \bibinfo{year}{2006}\natexlab{}.
\newblock \showarticletitle{A tutorial on energy-based learning}.
\newblock \bibinfo{journal}{\emph{Predicting structured data}} \bibinfo{volume}{1}, \bibinfo{number}{0} (\bibinfo{year}{2006}).
\newblock


\bibitem[Li et~al\mbox{.}(2024)]%
        {li2024dngaussian}
\bibfield{author}{\bibinfo{person}{Jiahe Li}, \bibinfo{person}{Jiawei Zhang}, \bibinfo{person}{Xiao Bai}, \bibinfo{person}{Jin Zheng}, \bibinfo{person}{Xin Ning}, \bibinfo{person}{Jun Zhou}, {and} \bibinfo{person}{Lin Gu}.} \bibinfo{year}{2024}\natexlab{}.
\newblock \showarticletitle{DNGaussian: Optimizing Sparse-View 3D Gaussian Radiance Fields with Global-Local Depth Normalization}. In \bibinfo{booktitle}{\emph{2024 IEEE/CVF Conference on Computer Vision and Pattern Recognition (CVPR)}}. \bibinfo{pages}{20775--20785}.
\newblock
\href{https://doi.org/10.1109/CVPR52733.2024.01963}{doi:\nolinkurl{10.1109/CVPR52733.2024.01963}}


\bibitem[Liu et~al\mbox{.}(2026)]%
        {liu2024reconx}
\bibfield{author}{\bibinfo{person}{Fangfu Liu}, \bibinfo{person}{Wenqiang Sun}, \bibinfo{person}{Hanyang Wang}, \bibinfo{person}{Yikai Wang}, \bibinfo{person}{Haowen Sun}, \bibinfo{person}{Junliang Ye}, \bibinfo{person}{Jun Zhang}, {and} \bibinfo{person}{Yueqi Duan}.} \bibinfo{year}{2026}\natexlab{}.
\newblock \showarticletitle{ReconX: Reconstruct Any Scene From Sparse Views With Video Diffusion Model}.
\newblock \bibinfo{journal}{\emph{IEEE Transactions on Image Processing}}  \bibinfo{volume}{35} (\bibinfo{year}{2026}), \bibinfo{pages}{2305--2319}.
\newblock
\href{https://doi.org/10.1109/TIP.2026.3666733}{doi:\nolinkurl{10.1109/TIP.2026.3666733}}


\bibitem[Matsuki et~al\mbox{.}(2024)]%
        {matsuki2024gaussian}
\bibfield{author}{\bibinfo{person}{Hidenobu Matsuki}, \bibinfo{person}{Riku Murai}, \bibinfo{person}{Paul~H.J. Kelly}, {and} \bibinfo{person}{Andrew~J. Davison}.} \bibinfo{year}{2024}\natexlab{}.
\newblock \showarticletitle{Gaussian Splatting SLAM}. In \bibinfo{booktitle}{\emph{Proceedings of the IEEE/CVF Conference on Computer Vision and Pattern Recognition (CVPR)}}. \bibinfo{pages}{18039--18048}.
\newblock


\bibitem[Mildenhall et~al\mbox{.}(2021)]%
        {mildenhall2021nerf}
\bibfield{author}{\bibinfo{person}{Ben Mildenhall}, \bibinfo{person}{Pratul~P. Srinivasan}, \bibinfo{person}{Matthew Tancik}, \bibinfo{person}{Jonathan~T. Barron}, \bibinfo{person}{Ravi Ramamoorthi}, {and} \bibinfo{person}{Ren Ng}.} \bibinfo{year}{2021}\natexlab{}.
\newblock \showarticletitle{NeRF: representing scenes as neural radiance fields for view synthesis}.
\newblock \bibinfo{journal}{\emph{Commun. ACM}} \bibinfo{volume}{65}, \bibinfo{number}{1} (\bibinfo{date}{Dec.} \bibinfo{year}{2021}), \bibinfo{pages}{99–106}.
\newblock
\showISSN{0001-0782}
\href{https://doi.org/10.1145/3503250}{doi:\nolinkurl{10.1145/3503250}}


\bibitem[M\"{u}ller et~al\mbox{.}(2022)]%
        {muller2022instant}
\bibfield{author}{\bibinfo{person}{Thomas M\"{u}ller}, \bibinfo{person}{Alex Evans}, \bibinfo{person}{Christoph Schied}, {and} \bibinfo{person}{Alexander Keller}.} \bibinfo{year}{2022}\natexlab{}.
\newblock \showarticletitle{Instant neural graphics primitives with a multiresolution hash encoding}.
\newblock \bibinfo{journal}{\emph{ACM Trans. Graph.}} \bibinfo{volume}{41}, \bibinfo{number}{4}, Article \bibinfo{articleno}{102} (\bibinfo{date}{July} \bibinfo{year}{2022}), \bibinfo{numpages}{15}~pages.
\newblock
\showISSN{0730-0301}
\href{https://doi.org/10.1145/3528223.3530127}{doi:\nolinkurl{10.1145/3528223.3530127}}


\bibitem[Ni et~al\mbox{.}(2025)]%
        {ni2025wonderturbo}
\bibfield{author}{\bibinfo{person}{Chaojun Ni}, \bibinfo{person}{Xiaofeng Wang}, \bibinfo{person}{Zheng Zhu}, \bibinfo{person}{Weijie Wang}, \bibinfo{person}{Haoyun Li}, \bibinfo{person}{Guosheng Zhao}, \bibinfo{person}{Jie Li}, \bibinfo{person}{Wenkang Qin}, \bibinfo{person}{Guan Huang}, {and} \bibinfo{person}{Wenjun Mei}.} \bibinfo{year}{2025}\natexlab{}.
\newblock \showarticletitle{WonderTurbo: Generating Interactive 3D World in 0.72 Seconds}. In \bibinfo{booktitle}{\emph{Proceedings of the IEEE/CVF International Conference on Computer Vision (ICCV)}}. \bibinfo{pages}{27423--27434}.
\newblock


\bibitem[Niemeyer et~al\mbox{.}(2022)]%
        {niemeyer2022regnerf}
\bibfield{author}{\bibinfo{person}{Michael Niemeyer}, \bibinfo{person}{Jonathan~T. Barron}, \bibinfo{person}{Ben Mildenhall}, \bibinfo{person}{Mehdi S.~M. Sajjadi}, \bibinfo{person}{Andreas Geiger}, {and} \bibinfo{person}{Noha Radwan}.} \bibinfo{year}{2022}\natexlab{}.
\newblock \showarticletitle{RegNeRF: Regularizing Neural Radiance Fields for View Synthesis from Sparse Inputs}. In \bibinfo{booktitle}{\emph{2022 IEEE/CVF Conference on Computer Vision and Pattern Recognition (CVPR)}}. \bibinfo{pages}{5470--5480}.
\newblock
\href{https://doi.org/10.1109/CVPR52688.2022.00540}{doi:\nolinkurl{10.1109/CVPR52688.2022.00540}}


\bibitem[Papantonakis et~al\mbox{.}(2024)]%
        {papantonakis2024reducing}
\bibfield{author}{\bibinfo{person}{Panagiotis Papantonakis}, \bibinfo{person}{Georgios Kopanas}, \bibinfo{person}{Bernhard Kerbl}, \bibinfo{person}{Alexandre Lanvin}, {and} \bibinfo{person}{George Drettakis}.} \bibinfo{year}{2024}\natexlab{}.
\newblock \showarticletitle{Reducing the Memory Footprint of 3D Gaussian Splatting}.
\newblock \bibinfo{journal}{\emph{Proc. ACM Comput. Graph. Interact. Tech.}} \bibinfo{volume}{7}, \bibinfo{number}{1}, Article \bibinfo{articleno}{16} (\bibinfo{date}{May} \bibinfo{year}{2024}), \bibinfo{numpages}{17}~pages.
\newblock
\href{https://doi.org/10.1145/3651282}{doi:\nolinkurl{10.1145/3651282}}


\bibitem[Peng et~al\mbox{.}(2024)]%
        {peng2024rtg}
\bibfield{author}{\bibinfo{person}{Zhexi Peng}, \bibinfo{person}{Tianjia Shao}, \bibinfo{person}{Yong Liu}, \bibinfo{person}{Jingke Zhou}, \bibinfo{person}{Yin Yang}, \bibinfo{person}{Jingdong Wang}, {and} \bibinfo{person}{Kun Zhou}.} \bibinfo{year}{2024}\natexlab{}.
\newblock \showarticletitle{RTG-SLAM: Real-time 3D Reconstruction at Scale using Gaussian Splatting}. In \bibinfo{booktitle}{\emph{ACM SIGGRAPH 2024 Conference Papers}} (Denver, CO, USA) \emph{(\bibinfo{series}{SIGGRAPH '24})}. \bibinfo{publisher}{Association for Computing Machinery}, \bibinfo{address}{New York, NY, USA}, Article \bibinfo{articleno}{30}, \bibinfo{numpages}{11}~pages.
\newblock
\showISBNx{9798400705250}
\href{https://doi.org/10.1145/3641519.3657455}{doi:\nolinkurl{10.1145/3641519.3657455}}


\bibitem[Peng et~al\mbox{.}(2025)]%
        {peng2025gaussian}
\bibfield{author}{\bibinfo{person}{Zhexi Peng}, \bibinfo{person}{Kun Zhou}, {and} \bibinfo{person}{Tianjia Shao}.} \bibinfo{year}{2025}\natexlab{}.
\newblock \showarticletitle{Gaussian-Plus-SDF SLAM: High-Fidelity 3D Reconstruction at 150+ fps}.
\newblock \bibinfo{journal}{\emph{Computational Visual Media}} \bibinfo{volume}{11}, \bibinfo{number}{6} (\bibinfo{year}{2025}), \bibinfo{pages}{1195--1208}.
\newblock
\href{https://doi.org/10.26599/CVM.2025.9450513}{doi:\nolinkurl{10.26599/CVM.2025.9450513}}


\bibitem[Ren et~al\mbox{.}(2024)]%
        {ren2024scube}
\bibfield{author}{\bibinfo{person}{Xuanchi Ren}, \bibinfo{person}{Yifan Lu}, \bibinfo{person}{Hanxue Liang}, \bibinfo{person}{Zhangjie Wu}, \bibinfo{person}{Huan Ling}, \bibinfo{person}{Mike Chen}, \bibinfo{person}{Sanja Fidler}, \bibinfo{person}{Francis Williams}, {and} \bibinfo{person}{Jiahui Huang}.} \bibinfo{year}{2024}\natexlab{}.
\newblock \showarticletitle{SCube: Instant Large-Scale Scene Reconstruction using VoxSplats}. In \bibinfo{booktitle}{\emph{Advances in Neural Information Processing Systems}}, \bibfield{editor}{\bibinfo{person}{A.~Globerson}, \bibinfo{person}{L.~Mackey}, \bibinfo{person}{D.~Belgrave}, \bibinfo{person}{A.~Fan}, \bibinfo{person}{U.~Paquet}, \bibinfo{person}{J.~Tomczak}, {and} \bibinfo{person}{C.~Zhang}} (Eds.), Vol.~\bibinfo{volume}{37}. \bibinfo{publisher}{Curran Associates, Inc.}, \bibinfo{pages}{97670--97698}.
\newblock
\href{https://doi.org/10.52202/079017-3099}{doi:\nolinkurl{10.52202/079017-3099}}


\bibitem[Roberts and Hanrahan(2016)]%
        {roberts2016generating}
\bibfield{author}{\bibinfo{person}{Mike Roberts} {and} \bibinfo{person}{Pat Hanrahan}.} \bibinfo{year}{2016}\natexlab{}.
\newblock \showarticletitle{Generating dynamically feasible trajectories for quadrotor cameras}.
\newblock \bibinfo{journal}{\emph{ACM Trans. Graph.}} \bibinfo{volume}{35}, \bibinfo{number}{4}, Article \bibinfo{articleno}{61} (\bibinfo{date}{July} \bibinfo{year}{2016}), \bibinfo{numpages}{11}~pages.
\newblock
\showISSN{0730-0301}
\href{https://doi.org/10.1145/2897824.2925980}{doi:\nolinkurl{10.1145/2897824.2925980}}


\bibitem[Roberts et~al\mbox{.}(2017)]%
        {roberts2017submodular}
\bibfield{author}{\bibinfo{person}{Mike Roberts}, \bibinfo{person}{Shital Shah}, \bibinfo{person}{Debadeepta Dey}, \bibinfo{person}{Anh Truong}, \bibinfo{person}{Sudipta Sinha}, \bibinfo{person}{Ashish Kapoor}, \bibinfo{person}{Pat Hanrahan}, {and} \bibinfo{person}{Neel Joshi}.} \bibinfo{year}{2017}\natexlab{}.
\newblock \showarticletitle{Submodular Trajectory Optimization for Aerial 3D Scanning}. In \bibinfo{booktitle}{\emph{2017 IEEE International Conference on Computer Vision (ICCV)}}. \bibinfo{pages}{5334--5343}.
\newblock
\href{https://doi.org/10.1109/ICCV.2017.569}{doi:\nolinkurl{10.1109/ICCV.2017.569}}


\bibitem[Roessle et~al\mbox{.}(2023)]%
        {roessle2023ganerf}
\bibfield{author}{\bibinfo{person}{Barbara Roessle}, \bibinfo{person}{Norman M\"{u}ller}, \bibinfo{person}{Lorenzo Porzi}, \bibinfo{person}{Samuel~Rota Bul\`{o}}, \bibinfo{person}{Peter Kontschieder}, {and} \bibinfo{person}{Matthias Niessner}.} \bibinfo{year}{2023}\natexlab{}.
\newblock \showarticletitle{GANeRF: Leveraging Discriminators to Optimize Neural Radiance Fields}.
\newblock \bibinfo{journal}{\emph{ACM Trans. Graph.}} \bibinfo{volume}{42}, \bibinfo{number}{6}, Article \bibinfo{articleno}{207} (\bibinfo{date}{Dec.} \bibinfo{year}{2023}), \bibinfo{numpages}{14}~pages.
\newblock
\showISSN{0730-0301}
\href{https://doi.org/10.1145/3618402}{doi:\nolinkurl{10.1145/3618402}}


\bibitem[Rombach et~al\mbox{.}(2022)]%
        {rombach2022high}
\bibfield{author}{\bibinfo{person}{Robin Rombach}, \bibinfo{person}{Andreas Blattmann}, \bibinfo{person}{Dominik Lorenz}, \bibinfo{person}{Patrick Esser}, {and} \bibinfo{person}{Bj\"orn Ommer}.} \bibinfo{year}{2022}\natexlab{}.
\newblock \showarticletitle{High-Resolution Image Synthesis With Latent Diffusion Models}. In \bibinfo{booktitle}{\emph{Proceedings of the IEEE/CVF Conference on Computer Vision and Pattern Recognition (CVPR)}}. \bibinfo{pages}{10684--10695}.
\newblock


\bibitem[Ronneberger et~al\mbox{.}(2015)]%
        {ronneberger2015u}
\bibfield{author}{\bibinfo{person}{Olaf Ronneberger}, \bibinfo{person}{Philipp Fischer}, {and} \bibinfo{person}{Thomas Brox}.} \bibinfo{year}{2015}\natexlab{}.
\newblock \showarticletitle{U-Net: Convolutional Networks for Biomedical Image Segmentation}. In \bibinfo{booktitle}{\emph{Medical Image Computing and Computer-Assisted Intervention -- MICCAI 2015}}, \bibfield{editor}{\bibinfo{person}{Nassir Navab}, \bibinfo{person}{Joachim Hornegger}, \bibinfo{person}{William~M. Wells}, {and} \bibinfo{person}{Alejandro~F. Frangi}} (Eds.). \bibinfo{publisher}{Springer International Publishing}, \bibinfo{address}{Cham}, \bibinfo{pages}{234--241}.
\newblock
\showISBNx{978-3-319-24574-4}


\bibitem[Sabour et~al\mbox{.}(2025)]%
        {sabour2025spotlesssplats}
\bibfield{author}{\bibinfo{person}{Sara Sabour}, \bibinfo{person}{Lily Goli}, \bibinfo{person}{George Kopanas}, \bibinfo{person}{Mark Matthews}, \bibinfo{person}{Dmitry Lagun}, \bibinfo{person}{Leonidas Guibas}, \bibinfo{person}{Alec Jacobson}, \bibinfo{person}{David Fleet}, {and} \bibinfo{person}{Andrea Tagliasacchi}.} \bibinfo{year}{2025}\natexlab{}.
\newblock \showarticletitle{SpotLessSplats: Ignoring Distractors in 3D Gaussian Splatting}.
\newblock \bibinfo{journal}{\emph{ACM Trans. Graph.}} \bibinfo{volume}{44}, \bibinfo{number}{2}, Article \bibinfo{articleno}{17} (\bibinfo{date}{April} \bibinfo{year}{2025}), \bibinfo{numpages}{11}~pages.
\newblock
\showISSN{0730-0301}
\href{https://doi.org/10.1145/3727143}{doi:\nolinkurl{10.1145/3727143}}


\bibitem[Saharia et~al\mbox{.}(2022)]%
        {saharia2022palette}
\bibfield{author}{\bibinfo{person}{Chitwan Saharia}, \bibinfo{person}{William Chan}, \bibinfo{person}{Huiwen Chang}, \bibinfo{person}{Chris Lee}, \bibinfo{person}{Jonathan Ho}, \bibinfo{person}{Tim Salimans}, \bibinfo{person}{David Fleet}, {and} \bibinfo{person}{Mohammad Norouzi}.} \bibinfo{year}{2022}\natexlab{}.
\newblock \showarticletitle{Palette: Image-to-Image Diffusion Models}. In \bibinfo{booktitle}{\emph{ACM SIGGRAPH 2022 Conference Proceedings}} (Vancouver, BC, Canada) \emph{(\bibinfo{series}{SIGGRAPH '22})}. \bibinfo{publisher}{Association for Computing Machinery}, \bibinfo{address}{New York, NY, USA}, Article \bibinfo{articleno}{15}, \bibinfo{numpages}{10}~pages.
\newblock
\showISBNx{9781450393379}
\href{https://doi.org/10.1145/3528233.3530757}{doi:\nolinkurl{10.1145/3528233.3530757}}


\bibitem[Sauer et~al\mbox{.}(2024)]%
        {sauer2024adversarial}
\bibfield{author}{\bibinfo{person}{Axel Sauer}, \bibinfo{person}{Dominik Lorenz}, \bibinfo{person}{Andreas Blattmann}, {and} \bibinfo{person}{Robin Rombach}.} \bibinfo{year}{2024}\natexlab{}.
\newblock \showarticletitle{Adversarial Diffusion Distillation}. In \bibinfo{booktitle}{\emph{Computer Vision – ECCV 2024: 18th European Conference, Milan, Italy, September 29–October 4, 2024, Proceedings, Part LXXXVI}} (Milan, Italy). \bibinfo{publisher}{Springer-Verlag}, \bibinfo{address}{Berlin, Heidelberg}, \bibinfo{pages}{87–103}.
\newblock
\showISBNx{978-3-031-73015-3}
\href{https://doi.org/10.1007/978-3-031-73016-0_6}{doi:\nolinkurl{10.1007/978-3-031-73016-0_6}}


\bibitem[Sch{\"o}nberger et~al\mbox{.}(2016)]%
        {schonberger2016pixelwise}
\bibfield{author}{\bibinfo{person}{Johannes~L. Sch{\"o}nberger}, \bibinfo{person}{Enliang Zheng}, \bibinfo{person}{Jan-Michael Frahm}, {and} \bibinfo{person}{Marc Pollefeys}.} \bibinfo{year}{2016}\natexlab{}.
\newblock \showarticletitle{Pixelwise View Selection for Unstructured Multi-View Stereo}. In \bibinfo{booktitle}{\emph{Computer Vision -- ECCV 2016}}, \bibfield{editor}{\bibinfo{person}{Bastian Leibe}, \bibinfo{person}{Jiri Matas}, \bibinfo{person}{Nicu Sebe}, {and} \bibinfo{person}{Max Welling}} (Eds.). \bibinfo{publisher}{Springer International Publishing}, \bibinfo{address}{Cham}, \bibinfo{pages}{501--518}.
\newblock
\showISBNx{978-3-319-46487-9}


\bibitem[Schönberger and Frahm(2016)]%
        {schonberger2016structure}
\bibfield{author}{\bibinfo{person}{Johannes~L. Schönberger} {and} \bibinfo{person}{Jan-Michael Frahm}.} \bibinfo{year}{2016}\natexlab{}.
\newblock \showarticletitle{Structure-from-Motion Revisited}. In \bibinfo{booktitle}{\emph{2016 IEEE Conference on Computer Vision and Pattern Recognition (CVPR)}}. \bibinfo{pages}{4104--4113}.
\newblock
\href{https://doi.org/10.1109/CVPR.2016.445}{doi:\nolinkurl{10.1109/CVPR.2016.445}}


\bibitem[Song et~al\mbox{.}(2021)]%
        {song2020score}
\bibfield{author}{\bibinfo{person}{Yang Song}, \bibinfo{person}{Jascha Sohl{-}Dickstein}, \bibinfo{person}{Diederik~P. Kingma}, \bibinfo{person}{Abhishek Kumar}, \bibinfo{person}{Stefano Ermon}, {and} \bibinfo{person}{Ben Poole}.} \bibinfo{year}{2021}\natexlab{}.
\newblock \showarticletitle{Score-Based Generative Modeling through Stochastic Differential Equations}. In \bibinfo{booktitle}{\emph{9th International Conference on Learning Representations, {ICLR} 2021, Virtual Event, Austria, May 3-7, 2021}}. \bibinfo{publisher}{OpenReview.net}.
\newblock
\urldef\tempurl%
\url{https://openreview.net/forum?id=PxTIG12RRHS}
\showURL{%
\tempurl}


\bibitem[Straub et~al\mbox{.}(2019)]%
        {straub2019replica}
\bibfield{author}{\bibinfo{person}{Julian Straub}, \bibinfo{person}{Thomas Whelan}, \bibinfo{person}{Lingni Ma}, \bibinfo{person}{Yufan Chen}, \bibinfo{person}{Erik Wijmans}, \bibinfo{person}{Simon Green}, \bibinfo{person}{Jakob~J Engel}, \bibinfo{person}{Raul Mur-Artal}, \bibinfo{person}{Carl Ren}, \bibinfo{person}{Shobhit Verma}, {et~al\mbox{.}}} \bibinfo{year}{2019}\natexlab{}.
\newblock \showarticletitle{The replica dataset: A digital replica of indoor spaces}.
\newblock \bibinfo{journal}{\emph{arXiv preprint arXiv:1906.05797}} (\bibinfo{year}{2019}).
\newblock


\bibitem[Toft et~al\mbox{.}(2020)]%
        {toft2020single}
\bibfield{author}{\bibinfo{person}{Carl Toft}, \bibinfo{person}{Daniyar Turmukhambetov}, \bibinfo{person}{Torsten Sattler}, \bibinfo{person}{Fredrik Kahl}, {and} \bibinfo{person}{Gabriel~J. Brostow}.} \bibinfo{year}{2020}\natexlab{}.
\newblock \showarticletitle{Single-Image Depth Prediction Makes Feature Matching Easier}. In \bibinfo{booktitle}{\emph{Computer Vision -- ECCV 2020}}, \bibfield{editor}{\bibinfo{person}{Andrea Vedaldi}, \bibinfo{person}{Horst Bischof}, \bibinfo{person}{Thomas Brox}, {and} \bibinfo{person}{Jan-Michael Frahm}} (Eds.). \bibinfo{publisher}{Springer International Publishing}, \bibinfo{address}{Cham}, \bibinfo{pages}{473--492}.
\newblock
\showISBNx{978-3-030-58517-4}


\bibitem[Wang et~al\mbox{.}(2025)]%
        {wang2025pi}
\bibfield{author}{\bibinfo{person}{Yifan Wang}, \bibinfo{person}{Jianjun Zhou}, \bibinfo{person}{Haoyi Zhu}, \bibinfo{person}{Wenzheng Chang}, \bibinfo{person}{Yang Zhou}, \bibinfo{person}{Zizun Li}, \bibinfo{person}{Junyi Chen}, \bibinfo{person}{Jiangmiao Pang}, \bibinfo{person}{Chunhua Shen}, {and} \bibinfo{person}{Tong He}.} \bibinfo{year}{2025}\natexlab{}.
\newblock \showarticletitle{${\pi}^{3}$:Scalable Permutation-Equivariant Visual Geometry Learning}.
\newblock \bibinfo{journal}{\emph{arXiv preprint arXiv:2507.13347}} (\bibinfo{year}{2025}).
\newblock


\bibitem[Wang et~al\mbox{.}(2004)]%
        {wang2004image}
\bibfield{author}{\bibinfo{person}{Zhou Wang}, \bibinfo{person}{A.C. Bovik}, \bibinfo{person}{H.R. Sheikh}, {and} \bibinfo{person}{E.P. Simoncelli}.} \bibinfo{year}{2004}\natexlab{}.
\newblock \showarticletitle{Image quality assessment: from error visibility to structural similarity}.
\newblock \bibinfo{journal}{\emph{IEEE Transactions on Image Processing}} \bibinfo{volume}{13}, \bibinfo{number}{4} (\bibinfo{year}{2004}), \bibinfo{pages}{600--612}.
\newblock
\href{https://doi.org/10.1109/TIP.2003.819861}{doi:\nolinkurl{10.1109/TIP.2003.819861}}


\bibitem[Wu et~al\mbox{.}(2025b)]%
        {wu2025difix3d+}
\bibfield{author}{\bibinfo{person}{Jay~Zhangjie Wu}, \bibinfo{person}{Yuxuan Zhang}, \bibinfo{person}{Haithem Turki}, \bibinfo{person}{Xuanchi Ren}, \bibinfo{person}{Jun Gao}, \bibinfo{person}{Mike~Zheng Shou}, \bibinfo{person}{Sanja Fidler}, \bibinfo{person}{Zan Gojcic}, {and} \bibinfo{person}{Huan Ling}.} \bibinfo{year}{2025}\natexlab{b}.
\newblock \showarticletitle{Difix3D+: Improving 3D Reconstructions with Single-Step Diffusion Models}. In \bibinfo{booktitle}{\emph{2025 IEEE/CVF Conference on Computer Vision and Pattern Recognition (CVPR)}}. \bibinfo{pages}{26024--26035}.
\newblock
\href{https://doi.org/10.1109/CVPR52734.2025.02424}{doi:\nolinkurl{10.1109/CVPR52734.2025.02424}}


\bibitem[Wu et~al\mbox{.}(2025a)]%
        {wu2025genfusion}
\bibfield{author}{\bibinfo{person}{Sibo Wu}, \bibinfo{person}{Congrong Xu}, \bibinfo{person}{Binbin Huang}, \bibinfo{person}{Andreas Geiger}, {and} \bibinfo{person}{Anpei Chen}.} \bibinfo{year}{2025}\natexlab{a}.
\newblock \showarticletitle{GenFusion: Closing the Loop between Reconstruction and Generation via Videos}. In \bibinfo{booktitle}{\emph{2025 IEEE/CVF Conference on Computer Vision and Pattern Recognition (CVPR)}}. \bibinfo{pages}{6078--6088}.
\newblock
\href{https://doi.org/10.1109/CVPR52734.2025.00570}{doi:\nolinkurl{10.1109/CVPR52734.2025.00570}}


\bibitem[Wu et~al\mbox{.}(2024)]%
        {wu2024recent}
\bibfield{author}{\bibinfo{person}{Tong Wu}, \bibinfo{person}{Yu-Jie Yuan}, \bibinfo{person}{Ling-Xiao Zhang}, \bibinfo{person}{Jie Yang}, \bibinfo{person}{Yan-Pei Cao}, \bibinfo{person}{Ling-Qi Yan}, {and} \bibinfo{person}{Lin Gao}.} \bibinfo{year}{2024}\natexlab{}.
\newblock \showarticletitle{Recent advances in 3D Gaussian splatting}.
\newblock \bibinfo{journal}{\emph{Computational Visual Media}} \bibinfo{volume}{10}, \bibinfo{number}{4} (\bibinfo{year}{2024}), \bibinfo{pages}{613--642}.
\newblock
\href{https://doi.org/10.1007/s41095-024-0436-y}{doi:\nolinkurl{10.1007/s41095-024-0436-y}}


\bibitem[Xie et~al\mbox{.}(2018)]%
        {xie2018creating}
\bibfield{author}{\bibinfo{person}{Ke Xie}, \bibinfo{person}{Hao Yang}, \bibinfo{person}{Shengqiu Huang}, \bibinfo{person}{Dani Lischinski}, \bibinfo{person}{Marc Christie}, \bibinfo{person}{Kai Xu}, \bibinfo{person}{Minglun Gong}, \bibinfo{person}{Daniel Cohen-Or}, {and} \bibinfo{person}{Hui Huang}.} \bibinfo{year}{2018}\natexlab{}.
\newblock \showarticletitle{Creating and chaining camera moves for quadrotor videography}.
\newblock \bibinfo{journal}{\emph{ACM Trans. Graph.}} \bibinfo{volume}{37}, \bibinfo{number}{4}, Article \bibinfo{articleno}{88} (\bibinfo{date}{July} \bibinfo{year}{2018}), \bibinfo{numpages}{13}~pages.
\newblock
\showISSN{0730-0301}
\href{https://doi.org/10.1145/3197517.3201284}{doi:\nolinkurl{10.1145/3197517.3201284}}


\bibitem[Yang et~al\mbox{.}(2024)]%
        {yang2024gaussianobject}
\bibfield{author}{\bibinfo{person}{Chen Yang}, \bibinfo{person}{Sikuang Li}, \bibinfo{person}{Jiemin Fang}, \bibinfo{person}{Ruofan Liang}, \bibinfo{person}{Lingxi Xie}, \bibinfo{person}{Xiaopeng Zhang}, \bibinfo{person}{Wei Shen}, {and} \bibinfo{person}{Qi Tian}.} \bibinfo{year}{2024}\natexlab{}.
\newblock \showarticletitle{GaussianObject: High-Quality 3D Object Reconstruction from Four Views with Gaussian Splatting}.
\newblock \bibinfo{journal}{\emph{ACM Trans. Graph.}} \bibinfo{volume}{43}, \bibinfo{number}{6}, Article \bibinfo{articleno}{199} (\bibinfo{date}{Nov.} \bibinfo{year}{2024}), \bibinfo{numpages}{13}~pages.
\newblock
\showISSN{0730-0301}
\href{https://doi.org/10.1145/3687759}{doi:\nolinkurl{10.1145/3687759}}


\bibitem[Yang et~al\mbox{.}(2018)]%
        {yang2018uncut}
\bibfield{author}{\bibinfo{person}{Hao Yang}, \bibinfo{person}{Ke Xie}, \bibinfo{person}{Shengqiu Huang}, {and} \bibinfo{person}{Hui Huang}.} \bibinfo{year}{2018}\natexlab{}.
\newblock \showarticletitle{Uncut Aerial Video via a Single Sketch}.
\newblock \bibinfo{journal}{\emph{Computer Graphics Forum}} \bibinfo{volume}{37}, \bibinfo{number}{7} (\bibinfo{year}{2018}), \bibinfo{pages}{191--199}.
\newblock
\showeprint{https://onlinelibrary.wiley.com/doi/pdf/10.1111/cgf.13559}
\href{https://doi.org/10.1111/cgf.13559}{doi:\nolinkurl{10.1111/cgf.13559}}


\bibitem[Yang et~al\mbox{.}(2023)]%
        {yang2023freenerf}
\bibfield{author}{\bibinfo{person}{Jiawei Yang}, \bibinfo{person}{Marco Pavone}, {and} \bibinfo{person}{Yue Wang}.} \bibinfo{year}{2023}\natexlab{}.
\newblock \showarticletitle{FreeNeRF: Improving Few-Shot Neural Rendering with Free Frequency Regularization}. In \bibinfo{booktitle}{\emph{2023 IEEE/CVF Conference on Computer Vision and Pattern Recognition (CVPR)}}. \bibinfo{pages}{8254--8263}.
\newblock
\href{https://doi.org/10.1109/CVPR52729.2023.00798}{doi:\nolinkurl{10.1109/CVPR52729.2023.00798}}


\bibitem[Ye et~al\mbox{.}(2025)]%
        {ye2025gaussian}
\bibfield{author}{\bibinfo{person}{Keyang Ye}, \bibinfo{person}{Tianjia Shao}, {and} \bibinfo{person}{Kun Zhou}.} \bibinfo{year}{2025}\natexlab{}.
\newblock \showarticletitle{When Gaussian Meets Surfel: Ultra-fast High-fidelity Radiance Field Rendering}.
\newblock \bibinfo{journal}{\emph{ACM Trans. Graph.}} \bibinfo{volume}{44}, \bibinfo{number}{4}, Article \bibinfo{articleno}{113} (\bibinfo{date}{July} \bibinfo{year}{2025}), \bibinfo{numpages}{15}~pages.
\newblock
\showISSN{0730-0301}
\href{https://doi.org/10.1145/3730925}{doi:\nolinkurl{10.1145/3730925}}


\bibitem[Yeshwanth et~al\mbox{.}(2023)]%
        {yeshwanth2023scannet++}
\bibfield{author}{\bibinfo{person}{Chandan Yeshwanth}, \bibinfo{person}{Yueh-Cheng Liu}, \bibinfo{person}{Matthias Nießner}, {and} \bibinfo{person}{Angela Dai}.} \bibinfo{year}{2023}\natexlab{}.
\newblock \showarticletitle{ScanNet++: A High-Fidelity Dataset of 3D Indoor Scenes}. In \bibinfo{booktitle}{\emph{2023 IEEE/CVF International Conference on Computer Vision (ICCV)}}. \bibinfo{pages}{12--22}.
\newblock
\href{https://doi.org/10.1109/ICCV51070.2023.00008}{doi:\nolinkurl{10.1109/ICCV51070.2023.00008}}


\bibitem[Yu et~al\mbox{.}(2021)]%
        {yu2021plenoctrees}
\bibfield{author}{\bibinfo{person}{Alex Yu}, \bibinfo{person}{Ruilong Li}, \bibinfo{person}{Matthew Tancik}, \bibinfo{person}{Hao Li}, \bibinfo{person}{Ren Ng}, {and} \bibinfo{person}{Angjoo Kanazawa}.} \bibinfo{year}{2021}\natexlab{}.
\newblock \showarticletitle{PlenOctrees for Real-time Rendering of Neural Radiance Fields}. In \bibinfo{booktitle}{\emph{2021 IEEE/CVF International Conference on Computer Vision (ICCV)}}. \bibinfo{pages}{5732--5741}.
\newblock
\href{https://doi.org/10.1109/ICCV48922.2021.00570}{doi:\nolinkurl{10.1109/ICCV48922.2021.00570}}


\bibitem[Yu et~al\mbox{.}(2025)]%
        {yu2025wonderworld}
\bibfield{author}{\bibinfo{person}{Hong-Xing Yu}, \bibinfo{person}{Haoyi Duan}, \bibinfo{person}{Charles Herrmann}, \bibinfo{person}{William~T. Freeman}, {and} \bibinfo{person}{Jiajun Wu}.} \bibinfo{year}{2025}\natexlab{}.
\newblock \showarticletitle{WonderWorld: Interactive 3D Scene Generation from a Single Image}. In \bibinfo{booktitle}{\emph{2025 IEEE/CVF Conference on Computer Vision and Pattern Recognition (CVPR)}}. \bibinfo{pages}{5916--5926}.
\newblock
\href{https://doi.org/10.1109/CVPR52734.2025.00555}{doi:\nolinkurl{10.1109/CVPR52734.2025.00555}}


\bibitem[Yu et~al\mbox{.}(2023)]%
        {yu2023freedom}
\bibfield{author}{\bibinfo{person}{Jiwen Yu}, \bibinfo{person}{Yinhuai Wang}, \bibinfo{person}{Chen Zhao}, \bibinfo{person}{Bernard Ghanem}, {and} \bibinfo{person}{Jian Zhang}.} \bibinfo{year}{2023}\natexlab{}.
\newblock \showarticletitle{FreeDoM: Training-Free Energy-Guided Conditional Diffusion Model}. In \bibinfo{booktitle}{\emph{2023 IEEE/CVF International Conference on Computer Vision (ICCV)}}. \bibinfo{pages}{23117--23127}.
\newblock
\href{https://doi.org/10.1109/ICCV51070.2023.02118}{doi:\nolinkurl{10.1109/ICCV51070.2023.02118}}


\bibitem[Yu et~al\mbox{.}(2024)]%
        {yu2024mip}
\bibfield{author}{\bibinfo{person}{Zehao Yu}, \bibinfo{person}{Anpei Chen}, \bibinfo{person}{Binbin Huang}, \bibinfo{person}{Torsten Sattler}, {and} \bibinfo{person}{Andreas Geiger}.} \bibinfo{year}{2024}\natexlab{}.
\newblock \showarticletitle{Mip-Splatting: Alias-Free 3D Gaussian Splatting}. In \bibinfo{booktitle}{\emph{2024 IEEE/CVF Conference on Computer Vision and Pattern Recognition (CVPR)}}. \bibinfo{pages}{19447--19456}.
\newblock
\href{https://doi.org/10.1109/CVPR52733.2024.01839}{doi:\nolinkurl{10.1109/CVPR52733.2024.01839}}


\bibitem[Zhang et~al\mbox{.}(2021)]%
        {zhang2021continuous}
\bibfield{author}{\bibinfo{person}{Han Zhang}, \bibinfo{person}{Yucong Yao}, \bibinfo{person}{Ke Xie}, \bibinfo{person}{Chi-Wing Fu}, \bibinfo{person}{Hao Zhang}, {and} \bibinfo{person}{Hui Huang}.} \bibinfo{year}{2021}\natexlab{}.
\newblock \showarticletitle{Continuous aerial path planning for 3D urban scene reconstruction}.
\newblock \bibinfo{journal}{\emph{ACM Trans. Graph.}} \bibinfo{volume}{40}, \bibinfo{number}{6}, Article \bibinfo{articleno}{225} (\bibinfo{date}{Dec.} \bibinfo{year}{2021}), \bibinfo{numpages}{15}~pages.
\newblock
\showISSN{0730-0301}
\href{https://doi.org/10.1145/3478513.3480483}{doi:\nolinkurl{10.1145/3478513.3480483}}


\bibitem[Zhang et~al\mbox{.}(2024)]%
        {zhang2024fregs}
\bibfield{author}{\bibinfo{person}{Jiahui Zhang}, \bibinfo{person}{Fangneng Zhan}, \bibinfo{person}{Muyu Xu}, \bibinfo{person}{Shijian Lu}, {and} \bibinfo{person}{Eric Xing}.} \bibinfo{year}{2024}\natexlab{}.
\newblock \showarticletitle{FreGS: 3D Gaussian Splatting with Progressive Frequency Regularization}. In \bibinfo{booktitle}{\emph{2024 IEEE/CVF Conference on Computer Vision and Pattern Recognition (CVPR)}}. \bibinfo{pages}{21424--21433}.
\newblock
\href{https://doi.org/10.1109/CVPR52733.2024.02024}{doi:\nolinkurl{10.1109/CVPR52733.2024.02024}}


\bibitem[Zhang et~al\mbox{.}(2018)]%
        {zhang2018unreasonable}
\bibfield{author}{\bibinfo{person}{Richard Zhang}, \bibinfo{person}{Phillip Isola}, \bibinfo{person}{Alexei~A. Efros}, \bibinfo{person}{Eli Shechtman}, {and} \bibinfo{person}{Oliver Wang}.} \bibinfo{year}{2018}\natexlab{}.
\newblock \showarticletitle{The Unreasonable Effectiveness of Deep Features as a Perceptual Metric}. In \bibinfo{booktitle}{\emph{2018 IEEE/CVF Conference on Computer Vision and Pattern Recognition}}. \bibinfo{pages}{586--595}.
\newblock
\href{https://doi.org/10.1109/CVPR.2018.00068}{doi:\nolinkurl{10.1109/CVPR.2018.00068}}


\bibitem[Zhang et~al\mbox{.}(2025)]%
        {zhang2025scene}
\bibfield{author}{\bibinfo{person}{Shengjun Zhang}, \bibinfo{person}{Jinzhao Li}, \bibinfo{person}{Xin Fei}, \bibinfo{person}{Hao Liu}, {and} \bibinfo{person}{Yueqi Duan}.} \bibinfo{year}{2025}\natexlab{}.
\newblock \showarticletitle{Scene Splatter: Momentum 3D Scene Generation from Single Image with Video Diffusion Model}. In \bibinfo{booktitle}{\emph{2025 IEEE/CVF Conference on Computer Vision and Pattern Recognition (CVPR)}}. \bibinfo{pages}{6089--6098}.
\newblock
\href{https://doi.org/10.1109/CVPR52734.2025.00571}{doi:\nolinkurl{10.1109/CVPR52734.2025.00571}}


\bibitem[Zhang et~al\mbox{.}(2023)]%
        {zhang2023sine}
\bibfield{author}{\bibinfo{person}{Zhixing Zhang}, \bibinfo{person}{Ligong Han}, \bibinfo{person}{Arnab Ghosh}, \bibinfo{person}{Dimitris~N. Metaxas}, {and} \bibinfo{person}{Jian Ren}.} \bibinfo{year}{2023}\natexlab{}.
\newblock \showarticletitle{SINE: SINgle Image Editing With Text-to-Image Diffusion Models}. In \bibinfo{booktitle}{\emph{Proceedings of the IEEE/CVF Conference on Computer Vision and Pattern Recognition (CVPR)}}. \bibinfo{pages}{6027--6037}.
\newblock


\bibitem[Zhao et~al\mbox{.}(2022)]%
        {zhao2022egsde}
\bibfield{author}{\bibinfo{person}{Min Zhao}, \bibinfo{person}{Fan Bao}, \bibinfo{person}{Chongxuan Li}, {and} \bibinfo{person}{Jun Zhu}.} \bibinfo{year}{2022}\natexlab{}.
\newblock \showarticletitle{EGSDE: unpaired image-to-image translation via energy-guided stochastic differential equations}. In \bibinfo{booktitle}{\emph{Proceedings of the 36th International Conference on Neural Information Processing Systems}} (New Orleans, LA, USA) \emph{(\bibinfo{series}{NIPS '22})}. \bibinfo{publisher}{Curran Associates Inc.}, \bibinfo{address}{Red Hook, NY, USA}, Article \bibinfo{articleno}{261}, \bibinfo{numpages}{15}~pages.
\newblock
\showISBNx{9781713871088}


\bibitem[Zheng et~al\mbox{.}(2025)]%
        {zheng2025nexusgs}
\bibfield{author}{\bibinfo{person}{Yulong Zheng}, \bibinfo{person}{Zicheng Jiang}, \bibinfo{person}{Shengfeng He}, \bibinfo{person}{Yandu Sun}, \bibinfo{person}{Junyu Dong}, \bibinfo{person}{Huaidong Zhang}, {and} \bibinfo{person}{Yong Du}.} \bibinfo{year}{2025}\natexlab{}.
\newblock \showarticletitle{NexusGS: Sparse View Synthesis with Epipolar Depth Priors in 3D Gaussian Splatting}. In \bibinfo{booktitle}{\emph{2025 IEEE/CVF Conference on Computer Vision and Pattern Recognition (CVPR)}}. \bibinfo{pages}{26800--26809}.
\newblock
\href{https://doi.org/10.1109/CVPR52734.2025.02496}{doi:\nolinkurl{10.1109/CVPR52734.2025.02496}}


\bibitem[Zhong et~al\mbox{.}(2025)]%
        {zhong2025taming}
\bibfield{author}{\bibinfo{person}{Yingji Zhong}, \bibinfo{person}{Zhihao Li}, \bibinfo{person}{Dave~Zhenyu Chen}, \bibinfo{person}{Lanqing Hong}, {and} \bibinfo{person}{Dan Xu}.} \bibinfo{year}{2025}\natexlab{}.
\newblock \showarticletitle{Taming Video Diffusion Prior with Scene-Grounding Guidance for 3D Gaussian Splatting from Sparse Inputs}. In \bibinfo{booktitle}{\emph{2025 IEEE/CVF Conference on Computer Vision and Pattern Recognition (CVPR)}}. \bibinfo{pages}{6133--6143}.
\newblock
\href{https://doi.org/10.1109/CVPR52734.2025.00575}{doi:\nolinkurl{10.1109/CVPR52734.2025.00575}}


\end{thebibliography}
\clearpage

\end{document}